\documentclass{article}

\usepackage{microtype}
\usepackage{graphicx}
\usepackage{booktabs} 
\usepackage{subfig} 
\usepackage{multirow}

\usepackage{hyperref}
\usepackage[dvipsnames]{xcolor}

\usepackage[accepted]{icml2024}

\usepackage{amsmath}
\usepackage{amssymb}
\usepackage{mathtools}
\usepackage{amsthm}

\usepackage[capitalize,noabbrev]{cleveref}
\usepackage{balance}

\theoremstyle{plain}
\newtheorem{theorem}{Theorem}[section]

\theoremstyle{definition}
\newtheorem{definition}[theorem]{Definition}

\theoremstyle{remark}

\DeclareMathOperator{\tr}{tr}
\DeclareMathOperator*{\argmax}{arg\,max}

\newcommand{\transpose}{\mathsf{T}}

\usepackage[textsize=tiny]{todonotes}

\icmltitlerunning{Lie Neurons: Adjoint-Equivariant Neural Networks for Semisimple Lie Algebras}

\begin{document}

\twocolumn[
\icmltitle{Lie Neurons: Adjoint-Equivariant Neural Networks for Semisimple Lie Algebras}



\icmlsetsymbol{equal}{*}

\begin{icmlauthorlist}
\icmlauthor{Tzu-Yuan Lin}{equal,um}
\icmlauthor{Minghan Zhu}{equal,um}
\icmlauthor{Maani Ghaffari}{um}
\end{icmlauthorlist}

\icmlaffiliation{um}{University of Michigan, Ann Arbor, MI, USA}

\icmlcorrespondingauthor{Tzu-Yuan Lin}{tzuyuan@umich.edu}

\icmlkeywords{Geometric Learning, Equivariance, Representation Learning, Lie Group}

\vskip 0.3in
]



\printAffiliationsAndNotice{\icmlEqualContribution} 

\begin{abstract}

This paper proposes an equivariant neural network that takes data in any finite-dimensional semi-simple Lie algebra as input. The corresponding group acts on the Lie algebra as adjoint operations, making our proposed network adjoint-equivariant. Our framework generalizes the Vector Neurons, a simple $\mathrm{SO}(3)$-equivariant network, from 3-D Euclidean space to Lie algebra spaces, building upon the invariance property of the Killing form. Furthermore, we propose novel Lie bracket layers and geometric channel mixing layers that extend the modeling capacity. 
Experiments are conducted for the $\mathfrak{so}(3)$, $\mathfrak{sl}(3)$, and $\mathfrak{sp}(4)$ Lie algebras on various tasks, including fitting equivariant and invariant functions, learning system dynamics, point cloud registration, and homography-based shape classification. Our proposed equivariant network shows wide applicability and competitive performance in various domains. 
\end{abstract}

\section{Introduction}
For geometric problems in control theory, robotics, computer vision and graphics, Lie group methods provide the machinery to study continuous symmetries inherent to the problem~\citep{murray1994mathematical,liu2010computational,lynch2017modern,barrau2017invariant,van2020equivariant,yang2021larnet,lin2022legged, ghaffari2022progress}. Lie algebras are vector spaces that locally preserve the group structure, enabling efficient computation~\citep{teng2022lie, lin2023proprioceptive}. The standard group representation (linear group action) on Lie algebras is given by conjugation or adjoint action~\citep{hall2013lie}.


The equivariance property preserves the symmetry group structure, often a Lie group, such that the feature map commutes with the group representation. 
An equivariant model, by construction, generalizes over the variations caused by the group actions. Therefore, it reduces the sampling complexity in learning and improves the robustness and transparency facing input variations.
Equivariant models have gained success in various domains, including but not limited to the modeling of molecules \citep{thomas2018tensor}, physical systems \citep{finzi2020generalizing}, social networks \citep{maron2018invariant}, images \citep{worrall2017harmonic}, and point clouds \citep{zhu20234d}. Convolutional neural networks (CNNs) are translation-equivariant, enabling stable image features regardless of the pixel positions in the image plane. 
Typical extensions include rotation \citep{cohen2017convolutional} and scale \citep{worrall2019deep} equivariance, while more general extensions are also explored \citep{macdonald2022enabling}. 

\begin{figure}[t]
    \centering
    \subfloat[]{
        \includegraphics[trim={0cm 0 1cm 1cm},clip,width=0.48\columnwidth]{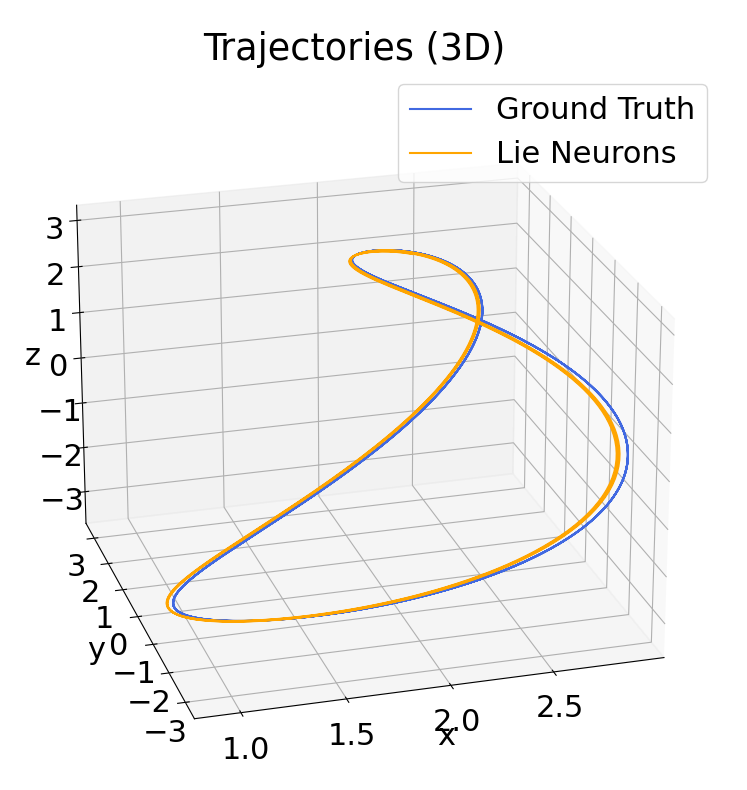}
        \label{fig:euler_poincare_3d_traj}}
    \subfloat[]{
        \includegraphics[trim={0cm 0 1cm 0cm},clip,width=0.48\columnwidth]{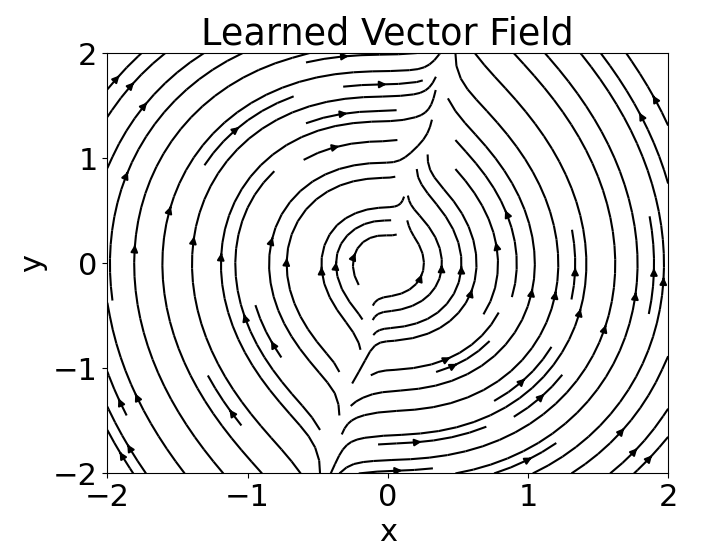}
        \label{fig:euler_poincare_vector_field}}\\
    \subfloat[]{
        \includegraphics[trim={3cm 0 3cm 1.5cm},clip,width=0.99\columnwidth]{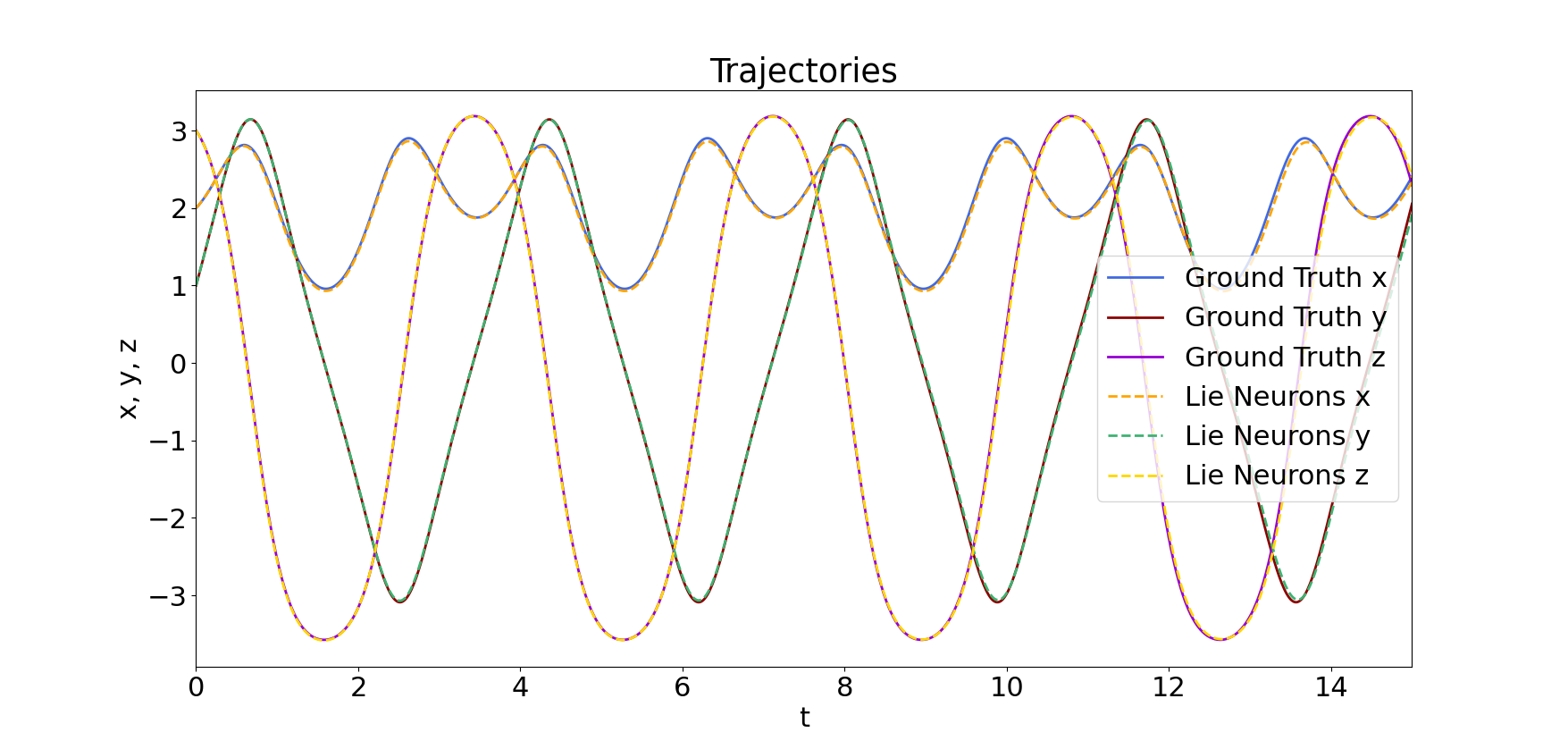}
        \label{fig:euler_poincare_2d_traj}}
    \caption{Lie Neurons can be applied in learning dynamics. In this example, we learn the dynamics of a simulated free-rotation International Space Station, which we pose as an initial value problem in the Neural ODE framework. This figure shows the estimated trajectories and the learned vector field from the Lie Neurons. Detail descriptions can be found in Section~\ref{sec:modeling_dynamics}.}
     \label{fig:euler_poincare_results}
     \vspace{-2mm}
\end{figure}

In this paper, we propose a new type of equivariant model that captures the symmetry in Lie algebra spaces. For a given Lie algebra, we represent its elements as vectors by specifying a set of bases, and represent the adjoint actions as matrix multiplications, exploiting the isometry structure of Lie algebras to handle such data in typical vector form. Through the connection between inner products and the Killing form, we generalize the architecture of Vector Neurons~\cite{deng2021vector}, a $\mathrm{SO}(3)$-equivariant network originally designed for point cloud data in 3-D Euclidean space, to process data in arbitrary semisimple Lie algebra. We further build new types of equivariant network layers that exploit the structure of Lie algebras and extend the flexibility of the model. Using $\mathfrak{so}(3)$, $\mathfrak{sl}(3)$, and $\mathfrak{sp}(4)$ as examples, we conduct experiments on various tasks in physics, computer vision, and general function fitting, and show that our proposed network has wide applicability and competitive performance. 

In particular, the contributions of this work are as follows.
\begin{enumerate}
    \item We propose a new adjoint-equivariant network architecture, enabling the processing of finite-dimensional semisimple Lie algebraic input data. Such models frequently appear in real-world geometric problems.
    \item We develop new network designs using the Killing form and the Lie bracket structure for equivariant activation and invariant layers for Lie algebraic representation learning. 
    \item We propose equivariant channel mixing layers that enable fusing information across the geometric dimension, which was not possible in the previous work~\cite{deng2021vector}. 
    \item The software implementation is available at \hyperlink{https://github.com/UMich-CURLY/LieNeurons}{https://github.com/UMich-CURLY/LieNeurons}. 
\end{enumerate}

\section{Related Work}
Equivariant networks enable the model output to change in a predicted way as the input goes through certain transformations. Current equivariant convolutional networks can be roughly categorized as regular group convolution and steerable group convolution. 
The regular group convolution exploits discretized subgroups to design equivariant convolutions. It is first introduced in \citet{cohen2016group}, in which 90-degree discretizations of the $SO(2)$ are designed for 2D image processing.
The approach is generalized to other discretized groups in $SE(2)$, $SE(3)$, $E(3)$, and $SIM(n)$ \citep{hoogeboom2018hexaconv, winkels20183d, worrall2018cubenet, chen2021equivariant, zhu2023e2pn, knigge2022exploiting}. Steerable convolution is proposed in \citet{cohen2016steerable}, leveraging the irreducible representations to remove the need for discretization and facilitate equivariant convolution on continuous groups in the frequency domain \citep{worrall2017harmonic, cohen2017convolutional, weiler20183d, thomas2018tensor}. Beyond convolutions, more general equivariant network architectures are proposed. For example, \citet{fuchs2020se, hutchinson2021lietransformer, chatzipantazis2022se} for transformers and \citet{batzner20223, brandstetter2021geometric} for message passing networks. More recently, \citet{geiger2022e3nn} introduces a generalized library for $E(3)$ equivariance based on variants of steerable equivariant networks. Vector Neurons \citep{deng2021vector} present a multi-layer perception (MLP) and graph network that generalize the scalar features to 3D features to realize $\mathrm{SO}(3)$-equivariance on spatial data. 
In addition to group-specific methods, more general recipes for building equivariant layers that are not limited to a specific group are also proposed \citep{kondor2018generalization, cohen2019general, weiler2019general, xu2022unified, lang2020wigner, bekkers2019b}. The extension of equivariance beyond compact groups is also explored. \citet{finzi2021practical} constructs MLPs equivariant to arbitrary matrix groups using their finite-dimensional representations. With the Monte Carlo estimator, equivariant convolutions are generalized to matrix groups with surjective exponential maps \citep{finzi2020generalizing} and all finite-dimensional Lie groups \citep{macdonald2022enabling}, where Lie algebras are used to parameterize elements in the continuous Lie groups as a lifted domain from the input space. More recently,~\citet{mironenco2023lie} proposes a theoretic framework for $SL(n)$-equivariant convolution by decomposing the larger groups into manageable subgroups. 

Our model structure resembles the MLP style of Vector Neurons \citep{deng2021vector}, but our work models the equivariance of arbitrary semisimple groups under adjoint actions. Different from Vector Neurons, the input domain of our method is the Lie algebra. When working with $\mathfrak{so}(3)$, our method specializes to Vector Neurons, with an additional nonlinearity and a geometric mixing layer. 

\section{Preliminaries}
\label{sec:preliminaries}
We provide some preliminaries for Lie groups by focusing on matrix Lie groups. For detailed explanations, we refer the readers to~\citet{hall2013lie,rossmann2006lie,kirillov2008introduction}.
\subsection{Lie Group and Lie Algebra}
A Lie group $\mathcal{G}$ is a smooth manifold whose elements satisfy the group axioms. 
The tangent space at the identity of a Lie group is named Lie algebra, denoted $\mathfrak{g}$.
A Lie algebra locally captures the structure of the Lie group. 

Every Lie algebra is equipped with an antisymmetric binary operator called the Lie bracket:
\begin{equation}
    \left[\cdot,\cdot\right]: \quad \mathfrak{g} \times \mathfrak{g} \rightarrow \mathfrak{g}.
\end{equation}

In this work, we focus on finite-dimensional Lie algebras. Since Lie algebra is a vector space, one can always find a set of basis $E_i \in \mathfrak{g}$, which are linearly independent matrices of the vector space. Once we find the basis, we can represent each Lie algebra element as a linear combination of such basis by collecting the coefficients of each basis into a $\mathbb{R}^m$ vector~\cite{chirikjian2011stochastic}:

\begin{align}
    \text{Vee}: & \mathfrak{g} \rightarrow \mathbb{R}^m, \quad x^{\wedge} \mapsto (x^{\wedge})^\vee = \sum^m_{i=1} x_i e_i\label{eq:veehat}, \quad \\
    \text{Hat}: & \mathbb{R}^m \rightarrow \mathfrak{g}, \quad x \mapsto x^{\wedge} = \sum^m_{i=1} x_i E_i,
\end{align}
where $E_i \in \mathfrak{g}$ are linear independent basis in $\mathfrak{g}$, and $e_i$ are the canonical basis of $\mathbb{R}^m$. An example of the Hat and Vee operation is provided in Appendix~\ref{sec:hat_vee_example}.

\subsection{Adjoint Representation}
Given an element of the Lie algebra $X \in \mathfrak{g}$ and its corresponding Lie group $\mathcal{G}$, every $a \in \mathcal{G}$ defines an automorphism of the Lie algebra $Ad_a : \mathfrak{g} \rightarrow \mathfrak{g}$ by
    $Ad_a(X) = a X a^{-1}$.
This is called the adjoint representation of the group $\mathcal{G}$ on the Lie algebra $\mathfrak{g}$. It amounts to the change of basis operations on the algebra. Since the adjoint $Ad_a$ is linear, we can find a matrix that maps the $\mathbb{R}^m$ form of the Lie algebra to another. That is, for every $Ad_a$ and $X \in \mathfrak{g}$, we have
\begin{equation}
    Adm_a: \quad \mathbb{R}^m \rightarrow \mathbb{R}^m, \quad x \mapsto Adm_a x,
    \label{eq:adjoint_matrix}
\end{equation}
with $Adm_a \in \mathbb{R}^{m\times m}$, $x = X^\vee$ and $Adm_a x = (a X a^{-1})^\vee$. This is an important property as it allows us to model the group adjoint action using a matrix multiplication on $\mathbb{R}^m$, which enables the adjoint equivariant layer design.


Similarly, we can obtain the adjoint representation of the Lie algebra as $ad_X : \mathfrak{g} \rightarrow \mathfrak{g}$ by $ad_X(Y) = \left[X,Y\right]$.
This work focuses on the matrix Lie group, where the Lie bracket is defined by the commutator: $\left[X,Y\right] = XY - YX$. It is worth noticing that the Lie bracket is equivariant under the group adjoint action, i.e., $\left[Ad_a(X),Ad_a(Y)\right] = Ad_a(\left[X,Y\right]), \forall a\in \mathcal{G}$. 

\subsection{Killing Form}
If a Lie algebra $\mathfrak{g}$ is of finite dimension and associated with a field $\mathbb{R}$, a symmetric bilinear form called the \textit{Killing form} is defined as~\citep{kirillov2008introduction}: 
\begin{equation}
    B(X,Y): \mathfrak{g} \times \mathfrak{g} \rightarrow \mathbb{R}, \quad (X,Y) \mapsto \tr(ad_X \circ ad_Y).
\end{equation}
\begin{definition}
    A bilinear form $B(X,Y)$ is said to be non-degenerate iff $B(X,Y) = 0$ for all $Y \in \mathfrak{g}$ implies $X = 0$.
\end{definition}
\begin{theorem}
    \label{theo:killing_non_degen}
    A Lie algebra is semisimple iff the Killing form is non-degenerate.\footnote{This is also known as the \textit{Cartan's Criterion}.}
\end{theorem}

\begin{theorem}
    \label{theo:killing_ad_inv}
    The Killing form is invariant under the group adjoint action $Ad_a$ for all $a \in \mathcal{G}$, i.e., \begin{equation*}
        B(Ad_a(X),Ad_a(Y)) = B(X,Y).
    \end{equation*}
\end{theorem}
If the Lie group is also compact, the Killing form is negative definite, and the inner product naturally arises from the negative of the Killing form.


\section{Methodology}
\label{sec:method}
We present Lie Neurons (LN), a general adjoint-equivariant neural network on Lie algebras. It is greatly inspired by Vector Neurons (VN)~\citep{deng2021vector}. Vector Neurons take 3-dimensional vectors as inputs, typically viewed as points in Euclidean space. The 3D Euclidean dimension is preserved in the features (which we call the \textit{geometric} dimension), independent from the \textit{feature} dimension. In other words, Vector Neurons lift conventional $\mathbb{R}^C$ features in an MLP to $\mathbb{R}^{3\times C}$, allowing the same $\mathrm{SO}(3)$ actions to be applied in the input space and the feature space, facilitating the equivariance property. 

Lie Neurons generalize VN with 3-channel geometric dimension to networks with $K$-channel geometric dimension, where $K$ is the dimension of any semisimple Lie algebra, and the networks are equivariant to the adjoint action of the corresponding Lie group. Similar to how VN takes 3-dimensional points as input, our networks take elements of the Lie algebra as input. While a Lie algebra $\mathfrak{g}$ is a vector space with non-trivial structures, $X\in\mathfrak{g}$ can be expressed as a $K$-dimensional vector $x = X^\vee \in \mathbb{R}^K$ with appropriate bases using~\eqref{eq:veehat}. This $\mathbb{R}^K$ form is the core concept for Lie Neurons.

Lie Neurons operate on the $\mathbb{R}^K$ form of the Lie algebra. We denote the input as $\mathbf{x}\in \mathbb{R}^{K \times C}$, where $C$ is the feature dimension. This can be viewed as $C$ Lie algebra elements in the $\mathbb{R}^K$ form. That is, a Lie Neurons model $f: \mathfrak{g}^{C_1} \rightarrow \mathfrak{g}^{C_2}$ can be instantiated as 
\begin{equation}
    f: \mathfrak{\mathbb{R}}^{K \times C_1} \rightarrow \mathfrak{\mathbb{R}}^{K \times C_2}.
\end{equation}

Recall from \eqref{eq:adjoint_matrix} that $Adm_a \in \mathbb{R}^{K\times K}$ is the matrix form of the adjoint operator such that $Adm_a x = (Ad_a(X))^\vee = (a X a^{-1})^\vee$. This means that we can represent the adjoint action as a left matrix multiplication on the input $\mathbf{x}$. 
The equivariance of the network can then be defined as \begin{equation}
    f(Adm_a \mathbf{x}; \theta) = Adm_a f(\mathbf{x}; \theta).
\end{equation}


For a set of $N$ input elements $\{X\}_N$, the learned feature is $\{\mathbf{x}\}_N\in \mathbb{R}^{K \times C\times N}$. We will use a single element to explain the network components for simplicity, unless noted otherwise. 

A Lie Neuron network can be constructed with linear layers, two types of nonlinear activation layers, geometric channel mixing layers, pooling layers, and invariant layers. 
We start by discussing the linear layers as follows.

\subsection{Linear Layers}
Linear layers are the basic building blocks of an MLP. A linear layer has a learnable weight matrix $W \in \mathbb{R}^{C \times C'}$, which operates on input features $\mathbf{x} \in \mathbb{R}^{K \times C}$ by right matrix multiplication:
\begin{equation}
    \mathbf{x}' = f_{\text{LN-Lin}}(\mathbf{x};W) = \mathbf{x}W \in \mathbb{R}^{K \times C' }.
\end{equation}
A linear layer can be viewed as a matrix multiplication on the right (feature dimension $C$), which does not affect the adjoint operation as matrix multiplication on the left (geometric dimension $K$), thus preserving the equivariance property:
\begin{align}
\begin{split}
    f_{\text{LN-Lin}}(Ad_a(\mathbf{x}); W) &= f_{\text{LN-Lin}}(Adm_a\mathbf{x}; W)\\
                                        &= Adm_a\mathbf{x}W \in \mathbb{R}^{K \times C'}\\
                                         &= Adm_a f_{\text{LN-Lin}}(\mathbf{x}; W)\\
                                         &= Ad_a(f_{\text{LN-Lin}}(\mathbf{x}; W)^\wedge),
\end{split}
\label{eq:linear_equi_proof}
\end{align}
It is worth mentioning that we ignore the bias term to preserve the equivariance. Lastly, similar to the Vector Neurons, the weights may or may not be shared across the elements $\mathbf{x}$ in $\{\mathbf{x}\}_N$.

\subsection{Nonlinear Layers}
Nonlinear layers enable the neural network to approximate complicated functions. We propose two designs for the equivariant nonlinear layers, \texttt{LN-ReLU} and \texttt{LN-Bracket}.  

\subsubsection{LN-ReLU: Nonlinearity Based on the Killing Form}
We can use an invariant function to construct an equivariant nonlinear layer. The VN leverages the inner product in a standard vector space, which is invariant to $\mathrm{SO(3)}$, to design a vector ReLU nonlinear layer. The idea is to "project" the negative part of a vector back to the zero plane, similar to how a 1-D ReLU rectifies the negative values. We generalize this idea by replacing the inner product with the negative of the Killing form. As described in \cref{sec:preliminaries}, the negative Killing form falls back to the inner product for compact semisimple Lie groups, and it is invariant to the group adjoint action. 

For an input $\mathbf{x} \in \mathbb{R}^{K \times C}$, a Killing form $B(\cdot,\cdot)$, and a learnable weight $U \in \mathbb{R}^{C \times C}$, the nonlinear layer $f_{\text{LN-ReLU}}$ is defined as:
\begin{equation}
    \label{eq:killing_relu}
    f_{\text{LN-ReLU}}(\mathbf{x}) =
    \begin{cases}
      \mathbf{x}, & \text{if}\ B(\mathbf{x},\mathbf{d}) \leq 0 \\
      \mathbf{x} + B(\mathbf{x},\mathbf{d})\mathbf{d}, & \text{otherwise},
    \end{cases}
\end{equation}
where $\mathbf{d} = \mathbf{x}U  \in \mathbb{R}^{K \times C}$ are learnable reference Lie algebra features. Optionally, one can also set $U \in \mathbb{R}^{C \times 1}$ so that $\mathbf{d}\in \mathbb{R}^{K \times 1}$, meaning that a single reference Lie algebra is shared across all $C$ feature channels. 

From \cref{theo:killing_ad_inv}, we know the Killing form is invariant under the group adjoint action, and the equivariance of the learned direction is proven in~\eqref{eq:linear_equi_proof}. Therefore, the second output of~\eqref{eq:killing_relu} becomes a linear combination of two equivariant quantities. As a result, the nonlinear layer is equivariant to the adjoint action.

We can also construct variants of ReLU, such as the leaky ReLU in the following form:
\begin{equation}
    f_{\text{LN-LeakyReLU}} = \alpha \mathbf{x} + (1-\alpha)f_{\text{LN-ReLU}}(\mathbf{x}) .
\end{equation}

\subsubsection{LN-Bracket: Nonlinearity Based on the Lie Bracket}
\label{sec:ln-bracket}
Lie algebra is a vector space with an extra binary operator called the Lie bracket, which is equivariant under group adjoint actions. Since we primarily focus on matrix groups, we use the commutator to build a novel nonlinear layer.

We use two learnable weight matrices $U, V\in \mathbb{R}^{C \times C}$ to map the input to different Lie algebra vectors, $\mathbf{u} = \mathbf{x}U, \mathbf{v}=\mathbf{x}V$. The Lie bracket of $\mathbf{u}$ and $\mathbf{v}$ becomes a nonlinear function on the input: $\mathbf{x} \mapsto [(\mathbf{x}U)^\wedge,(\mathbf{x}V)^\wedge]^\vee, \mathbb{R}^{K\times C}\rightarrow \mathbb{R}^{K\times C}$. Theoretically, we can use this as our nonlinear layer. However, we note that the Lie bracket essentially captures \textit{the failure of matrices to commute}~\citep{guggenheimer2012differential}, and that $\left[X,X\right] = 0, \forall X$. Thus, the bracket captures the “non-commutativity” created by the two linear maps, which can be small in practice. As a result, we add a skip connection to enhance the information flow, inspired by ResNet \citep{resnet2016}. The final design of the \texttt{LN-Bracket} layer becomes:
\begin{equation}
\label{eq:ln_bracket}
    f_{\text{LN-Bracket}}(\mathbf{x})= \mathbf{x} + [(\mathbf{x}U)^\wedge, (\mathbf{x}V)^\wedge]^\vee.
\end{equation}
The nonlinear layer is often combined with a linear layer to form a module. In the rest of the paper, we will use \texttt{LN-LR} to denote an \texttt{LN-Linear} followed by an \texttt{LN-ReLU}, and \texttt{LN-LB} to denote an \texttt{LN-Linear} with an \texttt{LN-Bracket} layer. 

\subsection{Geometric Channel Mixing}
\label{sec:channel_mixing}
One limitation of the Vector Neurons is the lack of a mixing mechanism in the geometric dimension $K$. In other words, the learnable weights are always multiplied on the right in the feature dimension, so that the equivariance to group actions as left matrix multiplications on the geometric dimension is preserved. However, mixing of the geometric channels is preferred or even necessary in some applications. (An example of such can be found in Section~\ref{sec:modeling_dynamics}.) Therefore, we proposed a geometric channel mixing mechanism for $\mathfrak{so}(n)$.

We construct the channel mixing module as:
\begin{equation}
    \mathbf{x}' = f_{\text{LN-Mix}}(\mathbf{x}) = \mathbf{M} \mathbf{x},
\end{equation}
where $\mathbf{M} = \mathbf{x}_1\mathbf{x}_2^\transpose \in \mathbb{R}^{K \times K}$, and
\begin{equation}
    \mathbf{x}_1, \mathbf{x}_2 = f_{\text{LN-ReLU}}(f_{\text{LN-Linear}}(\mathbf{x})) \in \mathbb{R}^{K \times C}
\end{equation}
are two learned equivariant features. One can easily verify the equivariance of the mixing module:
\begin{align}
\nonumber f_{\text{LN-Mix}}(R\mathbf{x}) &= R\mathbf{x}_1 \mathbf{x}_2^\transpose R^\transpose R \mathbf{x}
= R\mathbf{x}_1 \mathbf{x}_2^\transpose \mathbf{x} \\
&= R f_{\text{LN-Mix}}(\mathbf{x}),
\end{align}
where $R=Adm_R \in \mathrm{SO}(n)$.

This module enables information mixing in the geometric dimension for $\mathfrak{so}(n)$, which opens up the possibility to model functions with left multiplication on the inputs. Although the current mixing module only works for $\mathfrak{so}(n)$, we conjecture it is possible to extend to any semi-simple Lie algebra by $\mathbf{x}' = f_{\text{LN-Mix}}(\mathbf{x}) = \mathbf{x}_1 \mathbf{x}_2^{-1} \mathbf{x}$, where $\mathbf{x}_1, \mathbf{x}_2 \in \mathbb{R}^{K \times K}$. However, one needs to ensure the invertibility of $\mathbf{x}_2$, which may require post-processing and we leave for future discussion.

\subsection{Pooling Layers}\label{sec:pooling}
Pooling layers provide a means to aggregate global information across the $N$ input elements. This can be done by mean pooling, which is adjoint equivariant. In addition, we also introduce a max pooling layer. For input $\{\mathbf{x}_n \}_{n=1}^N \in \mathbb{R}^{K \times C \times N}$, and a weight matrix $W \in \mathbb{R}^{C \times C}$, we learn a set of directions as:
$\mathcal{D} = \{\mathbf{d}_n\}_{n=1}^{N} = \{\mathbf{x}_nW \}_{n=1}^{N}\in \mathbb{R}^{K \times C \times N}$.

We again employ the Killing form, $B(\cdot,\cdot)$, as the invariant function. For each feature channel $c \in C$, we have the max pooling function as $f_{\text{LN-Max}}(\mathbf{x}^c) = \mathbf{x}_{n^{*}}^c$, where
\begin{equation}
    n^{*}(c) = \argmax_n \ B(\mathbf{d}_n^c,\mathbf{x}_n^c),
\end{equation}
and $\mathbf{x}_n^c \in \mathbb{R}^K$ is the feature in $c^{th}$ channel of the $n^{th}$ element. 
Max pooling reduces the feature shape from $\mathbb{R}^{K \times C \times N}$ to $\mathbb{R}^{K \times C}$. The layer is equivariant to the adjoint action due to the invariance of $B(\cdot,\cdot)$.

\subsection{Invariant Layers}
Equivariant layers allow steerable feature learning. However, some applications demand invariant features~\cite{lin2023se,zheng2022rotation,li2021rotation}. We introduce an invariant layer that can be attached to the network when necessary. Given an input $\mathbf{x} \in \mathbb{R}^{K \times C}$, we have:
\begin{equation}
    f_{\text{LN-Inv}}(\mathbf{x}) = B(\mathbf{x},\mathbf{x}) \in \mathbb{R}^{C},
\end{equation}
where $B(\cdot,\cdot)$ is the adjoint-invariant Killing form.

\subsection{Relationship to Vector Neurons}
\label{sec:relat_vector_neurons}
Our method can specialize to the Vector Neurons when working with $\mathfrak{so}(3)$. This is because the linear adjoint matrix $Adm_a$ is exactly the rotation matrix for $\mathfrak{so}(3)$. Therefore, the group adjoint action becomes a left multiplication on the $\mathbb{R}^3$ form of the Lie algebra. Moreover, $\mathrm{SO}(3)$ is a compact group. Thus, the negative Killing form of $\mathfrak{so}(3)$ defines an inner product. However, we omit the normalization in VN's ReLU layer because the norm is not well defined when the Killing form is not negative definite. We also do not have a counterpart to VN's batch normalization layer for the same reason. 

Despite the similarity in appearance, the $\mathbb{R}^3$ vectors are viewed as points in the 3D Euclidean space in Vector Neurons, while they are treated as $\mathfrak{so}(3)$ Lie algebras in our framework, enabling a novel Lie bracket nonlinear layer. The geometric channel mixing layer also makes the model more flexible. 


\section{Experiments}
Our framework applies to arbitrary semi-simple Lie algebras. We conduct experiments on $\mathfrak{so}(3)$ and $\mathfrak{sl}(3)$ to validate its general applicability and effectiveness in various tasks. Implementation details and additional experiments on $\mathfrak{sl}(3)$ can be found in the Appendix.

\subsection{Experiment on $\mathfrak{sp}(4)$}
We conduct an experiment on the Symplectic Lie algebra $\mathfrak{sp}(4,\mathbb{R})$. The symplectic Lie algebra can be defined as: 
\begin{equation}
    \mathfrak{sp}(2n,\mathbb{R}) = \{ X \in GL(2n,\mathbb{R}) \mid MX+X^\intercal M = 0 \},
\end{equation} with $M = \begin{bmatrix} 0 &I_n \\ -I_n & 0\end{bmatrix}$. The symplectic group and algebra can be used to model Hamilton mechanics and energy-conservation systems.

In this task, we aim to regress an invariant function 
\begin{align}
    g(X,Y) &= \sin(Tr(XY))+\cos(Tr(YY))\\ \nonumber
           &-\frac{Tr(YY)^3}{2}+\det(XY)+\exp(Tr(XX)),
\end{align}

where $X,Y \in \mathfrak{sp}(4,\mathbb{R})$ and $Tr(\cdot)$ is the trace of the matrix. We generate 10,000 training and 10,000 testing data. In addition, during test time, we generate 500 $SP(4)$ matrices to perform test time augmentation. We report two MLP models with different feature dimensions for comparison. We also perform data augmentation during training for the MLP models. The invariance error captures the output consistency after the input is augmented. 

From Table~\ref{tab:sp4_inv_results}, we can see that Lie Neurons obtain superior accuracy while maintaining a low number of parameters compared to the MLP methods. In addition, the invariance error remains low without data augmentation, which confirms the invariant-by-construction property of Lie Neurons on the symplectic Lie algebra.

\begin{table}[t]
    \centering
    \caption{The mean squared errors and the invariance errors in $\mathfrak{sp}(4)$ invariant function regression.}
    \footnotesize
    \resizebox{\columnwidth}{!}{
    \begin{tabular}{c|c|c|c|c|c}
    \toprule
\multirow{3}{*}{Model} & \multirow{3}{*}{\begin{tabular}[c]{@{}c@{}}Training \\ Augmentation\end{tabular}} & \multirow{3}{*}{Num Params} & \multicolumn{2}{c|}{Testing Augmentation}                       & \multirow{2}{*}{Invariance Error} \\ \cmidrule{4-5} 
                       &                                                                                  &                             & $Id$       & $\mathrm{SP}(4)$    &                                  \\ \cmidrule{4-6}
                       &                                                                                   &                             & AVG     $\downarrow$               & AVG    $\downarrow$           & AVG    $\downarrow$                                         \\ \midrule
                    MLP 256	& Id	& 137,217	& 0.126	& 1.360	& 0.722\\
                    MLP 256	& SP(4)	& 137,217	& 0.192	& 0.587	& 0.476\\
                    MLP 512	& Id	& 536,577	& 0.107	& 0.906	& 0.585\\
                    MLP 512	& SP(4)	& 536,577	& 0.123	& 0.446	& 0.374\\
                    Lie Neurons	& Id & 263,170	& $\mathbf{2.70\times10^{-4}}$	& $\mathbf{2.70\times 10^{-4}}$ & $\mathbf{2.00\times10^{-4}}$\\
\bottomrule
\end{tabular}}
    \label{tab:sp4_inv_results}
    \vspace{-2mm}
\end{table}

\subsection{Experiments on $\mathfrak{so}(3)$}
As discussed in \cref{sec:relat_vector_neurons}, our network when applied on $\mathfrak{so}(3)$ resembles Vector Neurons \cite{deng2021vector}, but contains more flexible component layers. In this section, we present our results on the Baker–Campbell–Hausdorff (BCH) formula regression, dynamics learning, and point cloud registration. 


\subsubsection{Baker–Campbell–Hausdorff Formula}
The BCH formula provides a way to compute the product of two exponentials of elements in a Lie algebra locally~\cite{hall2013lie}. For $X, Y, Z \in \mathfrak{g}$, and
\begin{equation}
    e^Z = e^X e^Y,
    \label{eq:bch_exp}
\end{equation}
the BCH formula relates $Z$ to $X, Y$ in the Lie algebra by an infinite series:
\begin{align}
    Z &=  BCH(X,Y) \nonumber \\
    & = X + Y + \frac{1}{2} \left[X,Y\right] + \frac{1}{12} \left[X \left[X,Y\right] \right] + \cdots.
    \label{eq:bch_inf}
\end{align} We note that the BCH formula is adjoint equivariant. This formula is widely used in robotics and control~\cite{yoon2023invariant,chauchat2018invariant,kobilarov2011discrete}. However, the higher-order terms are often discarded in most applications, which can result in a significant drop in accuracy. An accurate modeling of the BCH formula can have great benefits in the above field.

We generate $10,000$ training and $10,000$ testing data points for both $X$ and $Y$ and use \eqref{eq:bch_exp} to generate the ground truth value. To ensure an injective exponential function, we limit the angle to $[0,\pi)$. We design the loss function to be: \begin{equation}
    L(X,Y\mid\theta) = \|e^Xe^Ye^{-f(X,Y)}-I\|_{\mathrm{F}},
\end{equation} where $\|\cdot\|_{\mathrm{F}}$ denotes the Frobenius norm, and $I$ is the identity matrix. The network architecture used in this experiment can be found in Figure~\ref{fig:ln_layers} in the Appendix. In addition to the Frobenius norm, we also report the log error, which is defined as: $E_{log} = ||\log(e^Xe^Ye^{-f(X,Y)})^\vee||,$ where $||\cdot||$ is the standard vector norm on $\mathbb{R}^3$.

The results are presented in Table~\ref{tab:bch_results}. We compare Lie Neurons with Vector Neurons~\cite{deng2021vector}, a steerable equivariant network e3nn~\cite{geiger2022e3nn}, the Equivariant MLP~\cite{finzi2021practical}, and an MLP with augmentation. In addition, we report the results obtained using \eqref{eq:bch_inf} by truncating to only first-third order terms. Both EMLP and our method achieve low estimation errors, while the error of our method is one order of magnitude smaller than EMLP. Although the MLP is able to produce acceptable results on the non-conjugated test set, its performance suffers from a significant drop when the inputs are adjoint-transformed.
With training augmentation, the performance of MLP is improved on the testing augmentation but is slightly reduced on the non-conjugated case. 
The e3nn and Vector Neurons are unable to converge in this experiment. Since Lie Neurons specialize to Vector Neurons on $\mathfrak{so}(3)$ if only the ReLU layer is used, this experiment demonstrates the benefits of the proposed bracket non-linear layer.

\begin{table}[t]
\centering
\caption{Experimental results on the regression of BCH formula. $Id$ represents testing on the original test set. $\mathrm{SO(3)}$ represents that the test set is augmented with adjoint actions. }
\resizebox{\columnwidth}{!}{
\begin{tabular}{l|cc|cc}
\toprule
\multirow{2}{*}{Method} & \multicolumn{2}{c|}{$Id$}                                       & \multicolumn{2}{c}{$\mathrm{SO(3)}$}                                    \\
                        & Frobenius Error  $\downarrow$              & Log Error   $\downarrow$                   & Frobenius Error     $\downarrow$           & Log Error      $\downarrow$                \\ \midrule
First Order Approx & 0.629 & 0.464 & 0.629 & 0.464\\
Second Order Approx & 0.338 & 0.247 & 0.338 & 0.247\\
Third Order Approx & 0.191 & 0.136 & 0.191 & 0.136\\
MLP                     & 0.017                          & 0.012                          & 0.295                          & 0.237                          \\
MLP Augmentation & 0.025 & 0.018 & 0.280 & 0.221\\
EMLP~\cite{finzi2021practical} & $2.6 \times 10^{-3}$ & $1.8 \times 10^{-3}$ & $2.6 \times 10^{-3}$ & $1.8 \times 10^{-3}$\\
e3nn~\cite{geiger2022e3nn} & 0.641 & 0.471 & 0.641 & 0.471\\
Vector Neurons~\cite{deng2021vector}          & 0.617 & 0.454 & 0.617 & 0.454  \\
Lie Neurons (\textit{Ours})             & $\mathbf{6.9\times 10^{-4}}$ & $\mathbf{4.9\times 10^{-4}}$ & $\mathbf{6.9\times 10^{-4}}$ & $\mathbf{4.9\times 10^{-4}}$ \\ \bottomrule
\end{tabular}
}\label{tab:bch_results}
\vspace{-2mm}
\end{table}

\subsubsection{Learning Dynamics in Body Frame}
\label{sec:modeling_dynamics}
Rigid body dynamics can be described using the Euler-Poincaré equation~\cite{bloch1996euler}. For a rotating rigid body, the Euler-Poincaré equation can be written as:
\begin{equation}
    \label{eq:euler_poincare}
    I \dot{\omega}(t) + \omega(t) \times I \omega(t) = M(t),
\end{equation}
where $\omega \in \mathbb{R}^3$ is the angular velocity, $I$ is the inertia tensor, and $M$ is the torque input. This equation is an ordinary differential equation (ODE). Given the ODE and an initial condition $\omega_0$ at time $t_0$, we can solve the initial value problem, i.e., predict the trajectory of the system. In this experiment, we learn the ODE from historic trajectory data and predict the trajectories of the learned system given arbitrary initial conditions. 

As a case study, we aim to learn the dynamics of the free-rotating (i.e., $M(t) = 0$) International Space Station (ISS) from~\citet{de2008orbit}. Specifically, we can rewrite the ODE as the vector field $\dot{\omega} = f(\omega; I)$ to represent the corresponding system dynamics, and we fit $f$ using a neural network. 
When a change of reference frame is performed, the inertia tensor undergoes a conjugation action. As a result, $f$ is equivariant under the change of reference frame: $\forall R\in \mathrm{SO}(3)$,
\begin{equation}
    f(R\omega; RIR^\transpose) = Rf(\omega; I).
\end{equation}
We introduce learnable equivariant weights $m \in \mathbb{R}^{3 \times C}$ as an implicit representation of the inertia $I$. 
When testing for the change of reference frame performance, we manually rotate $m$ to inform the network that the system is rotated.\footnote{It is possible to infer the change of reference frame matrix from observations of trajectories of the new system via another Lie Neurons module. Due to the scope of this project, we assume the change of frame matrix is known in test time.} For the network structure, please refer to \cref{fig:ln_layers} in the Appendix. 

\begin{figure}[t]
    \centering
    \includegraphics[width=0.8\columnwidth]{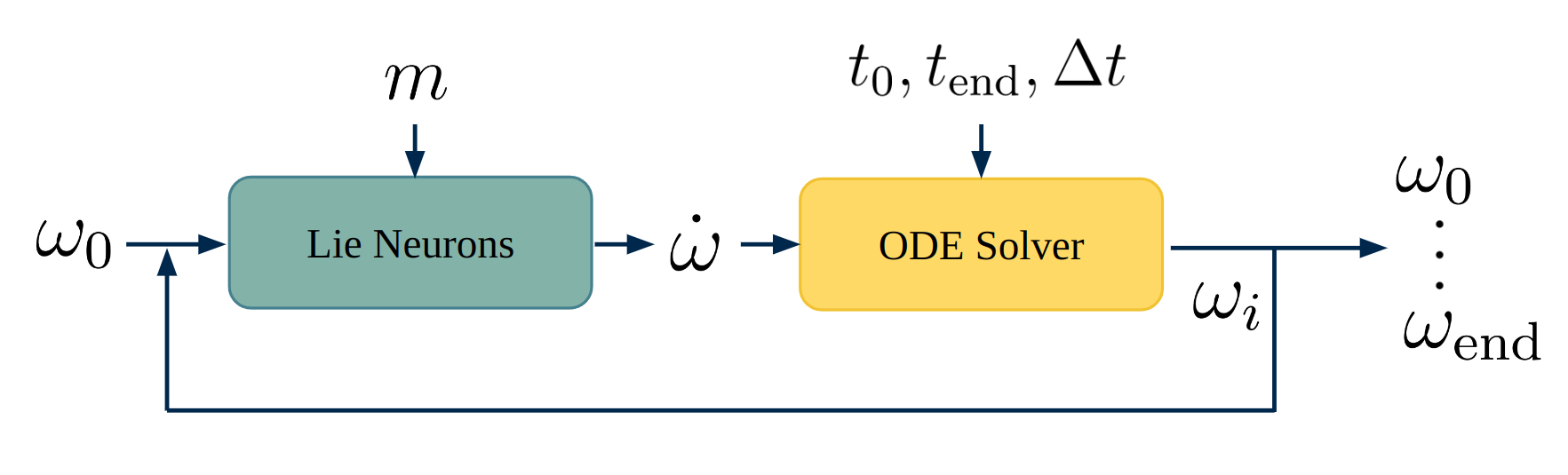}
    \caption{The framework used in modeling the Euler-Poincaré equation. Lie neurons learn the dynamic equation, and an off-the-shelf ODE solver is employed to solve the ODE. Here, $m$ is a set of learnable equivariant weights.}
    \label{fig:ln_ode_structure}
    \vspace{-2mm}
\end{figure}

We use the Neural ODE~\cite{chen2018neural} framework to train the ODE from trajectory data. 
As shown in \cref{fig:ln_ode_structure}, it consists of a neural network that models $f$ and an ODE solver. We compare our models with the model used in the original Neural ODE paper, which is an MLP network. In addition, we replace the MLP network with EMLP~\cite{finzi2021practical} to serve as a equivariant baseline.

\begin{table*}[t]
    \vspace{-5pt}
    \centering
    \scriptsize
    \caption{The results of the dynamic modeling experiments. We report the norm distance between the ground truth and estimated trajectories in different time durations. Multiple trajectories denote training on multiple random trajectories and evaluating using unseen data. Single trajectory denotes training and testing on the same trajectory. Unit: $\texttt{rad/s}$.}
    \begin{tabular}{l | c c c c c |c c c c c}
    \toprule
        \textbf{Multiple Trajectories} & \multicolumn{5}{c|}{$Id$} & \multicolumn{5}{c}{$\mathrm{SO(3)}$}\\
         \midrule
         Time (Sec) & 5 & 10 & 15 & 20 & 25 & 5 & 10 & 15 & 20 & 25 \\
         \midrule
         MLP & 0.428& 0.656 & 0.717 & 0.763 & 0.800 & 0.474 & 0.689 & 0.733 & 0.768 & 0.805\\
         EMLP~\cite{finzi2021practical} & 0.429 & 0.642 & 0.775 & 0.909 & 1.027 &	0.415 & 0.633 & 0.771 & 0.907 & 1.025\\
         Lie Neurons (No Mixing) & 0.739 & 0.842 & 0.791 & 0.805 & 0.809 & 0.739 & 0.842 & 0.791 & 0.805 & 0.809\\
         Lie Neurons &\textbf{0.005}& \textbf{0.011} &\textbf{0.014} & \textbf{0.016} & \textbf{0.018} &  \textbf{0.005} & \textbf{0.011} & \textbf{0.014} & \textbf{0.016} & \textbf{0.018} \\
         \midrule
             \textbf{Single Trajectory}  & \multicolumn{5}{c|}{$Id$} & \multicolumn{5}{c}{$\mathrm{SO(3)}$}\\
         \midrule
         Time (Sec) & 5 & 10 & 15 & 20 & 25 & 5 & 10 & 15 & 20 & 25 \\
         \midrule
         MLP & 0.108 & 0.137 & 0.162 & 0.200 & 0.225& 3.751 & 4.188 & 4.135 & 4.137 & 4.130\\
         EMLP~\cite{finzi2021practical} & 0.113 & 0.124 & \textbf{0.134} & \textbf{0.145} & \textbf{0.147} & 0.332 & 0.571 & 0.846 & 1.157 & 1.459\\
         Lie Neurons (No Mixing) & 0.720 & 0.824 & 0.801 & 0.821 & 0.836 & 0.720 & 0.824 & 0.801 & 0.821 & 0.836\\
         Lie Neurons &\textbf{0.064} & \textbf{0.069} & 0.146 & 0.324 & 0.579 &  \textbf{0.064} & \textbf{0.069} & \textbf{0.146} & \textbf{0.323} & \textbf{0.579} \\
        \bottomrule
    \end{tabular}
    \label{tab:euler_poincare_results}
    \vspace{-2mm}
\end{table*}
Using the inertia tensor from~\citet{de2008orbit}, we randomly generate $10$ trajectories, each containing $25$ seconds of data and $1000$ data points. We then evaluate the trained model using $10$ unseen trajectories in the test set. To analyze the equivariance property, we rotate the test inputs and trajectories using $10$ random rotations, which we denote as $\mathrm{SO(3)}$ in Table~\ref{tab:euler_poincare_results}. The table reports the norm distance between the estimated and the ground truth trajectories, evaluated at different points in time. 
We observe that the baselines are unable to predict the trajectories on the test set correctly. Therefore, we additionally report the results when both models are trained and evaluated on a single trajectory (The training and test sets are identical). 

The proposed method obtains accurate predictions for all experiments. The prediction error remains low when the reference frames are changed. In comparison, both the MLP and EMLP are unable to correctly predict the trajectories when evaluated on the unseen data. When trained and tested on the same trajectory, both the MLP and EMLP can overfit the data. However, the MLP fails to generalize when the reference frame is rotated, while the EMLP slightly alleviates such a problem.

We observe that when the mixing layers are not added, the network is unable to converge. It is likely because, in \eqref{eq:euler_poincare}, the inertia tensor acts on the left of the angular velocity, which results in mixing in the geometric dimension. Without the mixing layers, such an operation might not be modeled correctly, which demonstrates the value of the proposed mixing layers.

Figure~\ref{fig:euler_poincare_results} shows the qualitative results from Lie Neurons. The estimated trajectories closely align with the ground truth trajectory. Figure~\ref{fig:euler_poincare_vector_field} visualizes the learned vector field on the $\omega_z=0$ plane. This experiment demonstrates Lie Neurons' ability to model equivariant dynamical systems, which implies its potential in robotics applications.

\subsubsection{Point Cloud Registration}
In this experiment, we follow the setup in~\citep{zhu2022correspondence} and implement Lie Neurons in the point cloud registration task. The inputs are two noisy point clouds with different orientations. The goal of the network is to regress an $SO(3)$ rotation that best aligns the two point clouds in a correspondence-free manner. The network is trained and evaluated on ModelNet40, which contains 3D models of objects in 40 categories. 

We compare Lie Neurons with the mixing module against Vector Neuron. In addition, to demonstrate the benefits of geometric mixing, we integrate the mixing module in the original Vector Neuron to serve as an additional baseline.

Table~\ref{tab:pcl_reg} shows the average registration error in degrees. On average, the mixing module improves the performance of Vector Neurons. Lie Neurons perform similarly to Vector Neurons with the mixing module. This experiment once again demonstrates the benefits of the geometric mixing layer.

\subsection{Experiments on $\mathfrak{sl}(3)$}
In this section, we instantiate the LN on a noncompact Lie algebra, $\mathfrak{sl}(3)$, the special linear Lie algebra. $\mathfrak{sl}(3)$ can be represented using traceless matrices. The corresponding special linear group $\mathrm{SL}(3)$ can be represented using matrices with unit determinants. $\mathrm{SL}(3)$ has 8 degrees of freedom and can be used to model the homography transformation between images~\cite{hua2020nonlinear,zhan2022warped}.

We perform three experiments for $\mathfrak{sl}(3)$: a classification of Platonic solids and 2 function regression tasks. We report the function regression tasks in Appendix~\ref{sec:additional_sl3_exp}.


\begin{table}[t]
    \vspace{-5pt}
    \centering
    \scriptsize 
    \caption{The average registration error in degrees. Mixing denotes the addition of the geometric channel mixing module.}
    \begin{tabular}{c|c}
    \toprule
Model & Average Registration Error ($\deg$) $\downarrow$   
\\ \midrule
Vector Neurons        &   2.227  \\
Vector Neurons (Mixing) &   1.934  \\
Lie Neurons    (Mixing) &  \textbf{1.879}
\\ \bottomrule
\end{tabular}
    \label{tab:pcl_reg}
    \vspace{-2mm}
\end{table}

\subsubsection{Platonic Solid Classification}
The task is to classify polyhedrons from their projection on an image plane. 
While rotation equivariance naturally emerges for the 3D shape, the rotation equivariance relation is lost in the 2D projection of the 3D polyhedrons. Instead, the projection yields homography relations, which can be modeled using the $\mathrm{SL}(3)$ group~\cite{hua2020nonlinear,zhan2022warped}. When projected onto an image plane, the two neighboring faces of a polyhedron can be described using homography transformations, which are different for each polyhedron type. Therefore, we use the homography transforms among the projected neighboring faces as the input for polyhedron classification. 

Without loss of generality, we assume the camera intrinsic matrix $K$ to be identity. 
In this case, given a homography matrix $H\in \mathrm{SL}(3)$ that maps one face to another in the image plane, the homography between these two faces becomes $RHR^{-1}$ when we rotate the camera by $R \in \mathrm{SO}(3)\subset \mathrm{SL}(3)$.

\begin{figure}[t]
    \centering
    \subfloat[Tetrahedron]{
        \includegraphics[width=0.24\columnwidth]{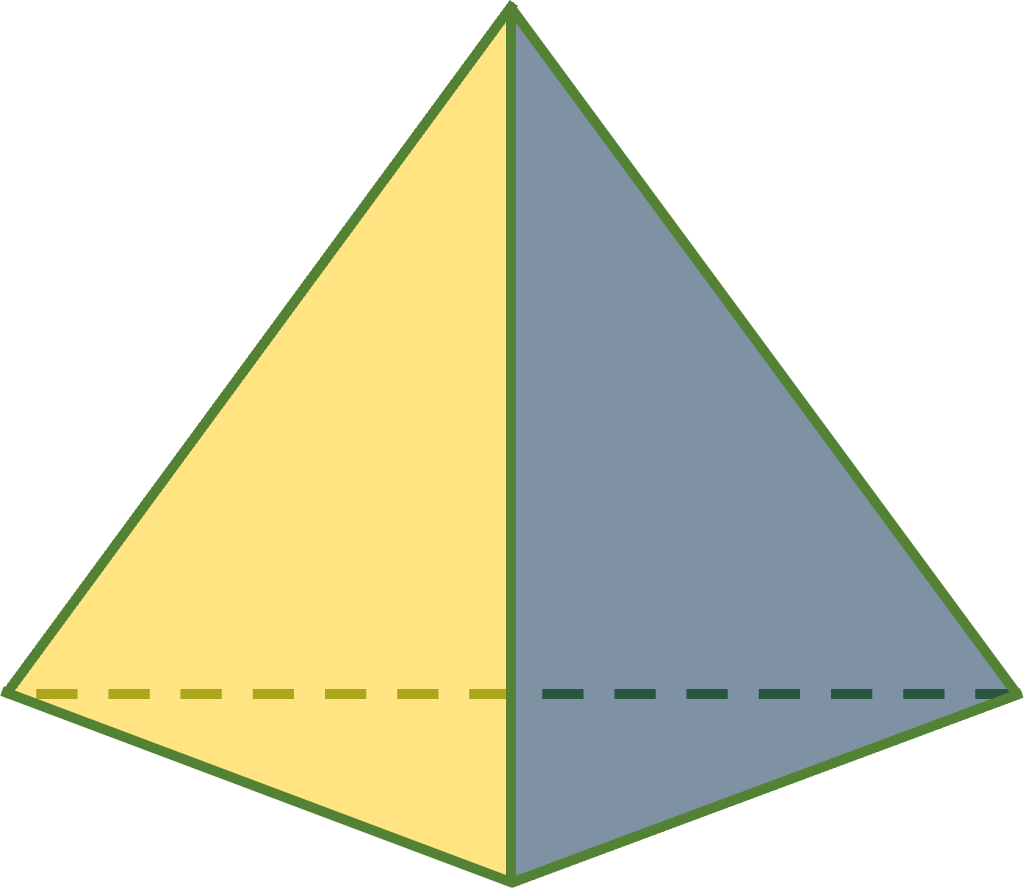}
        \label{fig:tetrahedron_ygb}}\hfill
    \subfloat[Octahedron]{
        \includegraphics[width=0.24\columnwidth]{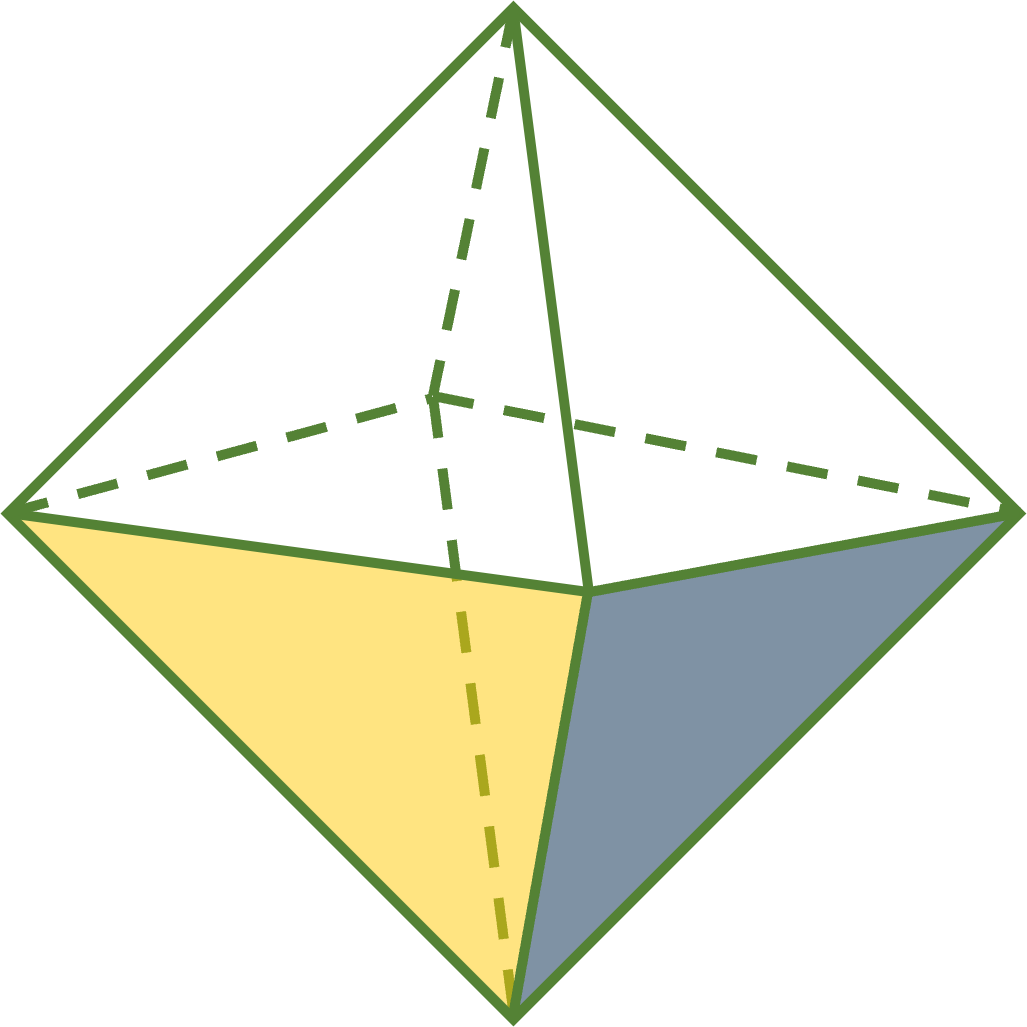}
        \label{fig:octahedron_ygb}}\hfill
    \subfloat[Icosahedron]{
        \includegraphics[width=0.22\columnwidth]{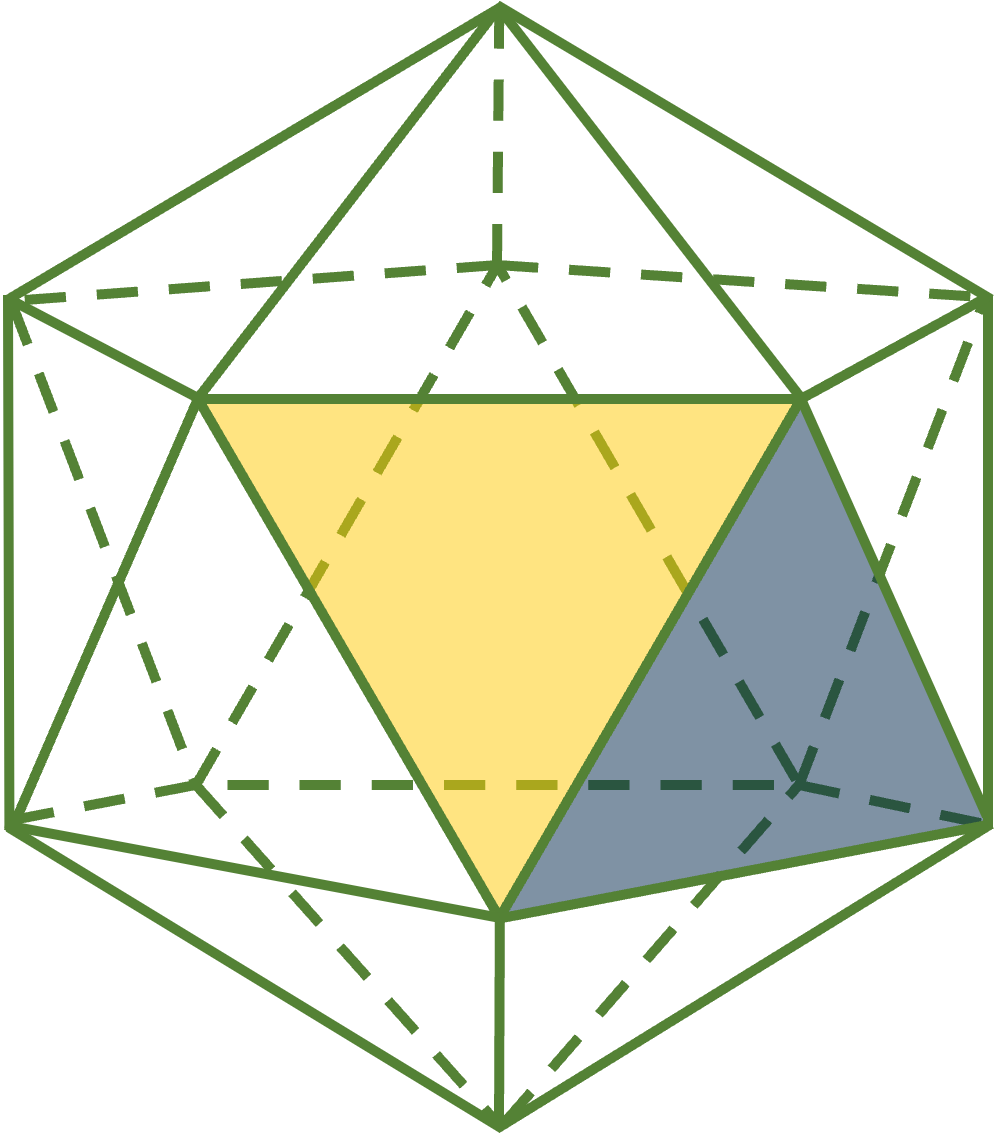}
        \label{fig:icosahedron_ygb}}
     \caption{A visualization of the three Platonic solids in our classification task. The yellow and blue colors highlight a neighboring pair of faces, between which the homography transforms in the image plane are taken as input to our models. }
     \label{fig:platonic}
     \vspace{-2mm}
\end{figure}

Three types of Platonic solids are used in this experiment: a tetrahedron, an octahedron, and an icosahedron. An input data point refers to the homography transforms between the projection of a pair of neighboring faces within one image. \cref{fig:platonic} visualizes an example of the neighboring face pair for the three Platonic solids. The homographies of all neighboring face pairs form a complete set of data describing a Platonic solid. We use these data to learn a classification model of the three Platonic solids. During training, we fix the camera and object pose. Then, we test with the original pose and with rotated camera poses to verify the equivariance property of our models. 


Each network is trained in 5 separate instances to analyze the consistency of the method. For the detailed architecture, we again refer the readers to Figure~\ref{fig:ln_layers}. Table~\ref{tab:classification_results} shows the classification accuracy. The LN achieves higher accuracy than the MLP. Since the MLP is not invariant to the adjoint action, its accuracy drops drastically when the camera is rotated. When trained with augmented data, the MLP performance on the rotated test set increases, but the overall performance decreases. We also notice that the \texttt{LN-LB} performs slightly worse than the other two formulations. 

\begin{table}[t]
    \vspace{-5pt}
    \centering
    \caption{The accuracy of the Platonic solid classification task using the inter-face homography transforms in the image plane as inputs. $\uparrow$ means the higher, the better.}
    \resizebox{\columnwidth}{!}{
    \begin{tabular}{cccccc}
    \toprule
Model   & Num Params  & \multicolumn{2}{c}{Acc $\uparrow$}                                    & \multicolumn{2}{c}{Acc (Rotated) $\uparrow$}                          \\
   &  & AVG & STD & AVG & STD \\ \midrule
MLP           & 206,339              & 95.76\%                          & 0.65\%                           & 36.54\%                          & 0.99\%                           \\
MLP Augmentation	& 206,339	& 81.47\% &	0.77\% &	81.20\%	& 2.34\% \\
LN-LR         & 134,664              & 99.56\%                          & 0.23\%                           & 99.51\%                          & 0.28\%                           \\
LN-LB         & 200,200              & 99.14\%                          & 0.21\%                           & 98.78\%                          & 0.49\%                           \\
LN-LR + LN-LB & 331,272              & \textbf{99.62}\%                          & 0.25\%                           & \textbf{99.61}\%                          & 0.14\%                          \\ \bottomrule
\end{tabular}}
    \label{tab:classification_results}
    \vspace{-2mm}
\end{table}

\section{Discussion and Limitations}
Lie Neurons are a group adjoint equivariant network by construction. It does not require the Lie group to be compact. However, the \texttt{LN-ReLU} layer relies on a non-degenerated Killing form, which limits the operation on semisimple Lie algebras for such a layer. For general Lie groups, the adjoint representation might not be irreducible. As a result, the linear layer may not cover all equivariant maps. However, given the complex representation theory of semisimple groups, we use the current linear layer design and increase flexibility with various nonlinear layers. The current mixing module only works for $\mathfrak{so}(n)$. Nevertheless, we conjecture it is possible to extend to any semi-simple Lie algebra, as discussed in Section~\ref{sec:channel_mixing}. Lie Neurons take elements in the Lie algebra as inputs, but most modern sensors return measurements in standard vector spaces. Finding equivariant lifts from the measurement space to the Lie algebraic space is an important future work. Lastly, this work assumes a basis can be found for the target Lie algebra, which is valid for many robotics and computer vision applications.

\section{Conclusion}
In this paper, we propose an adjoint-equivariant network, Lie Neurons, that models functions of Lie algebra elements. 
Our model is generally applicable to any semisimple Lie groups, compact or non-compact. 
Generalizing the Vector Neurons architecture, our network possesses simple MLP-style layers and can be viewed as a Lie algebraic extension to the MLP. 
To facilitate the learning of expressive Lie algebraic features, we propose equivariant nonlinear activation functions based on the Killing form and the Lie bracket. 
We also design an equivariant pooling layer and an invariant layer to extract global equivariant features and invariant features. Furthermore, a geometric mixing layer is proposed to facilitate information mixing in the geometric dimension, which was not possible in previous related work.

We demonstrate the effectiveness of the proposed method on several applications, including function regression on $\mathfrak{sp}(4)$, regression of the BCH formula on $\mathfrak{so}(3)$, dynamic modeling of the free-rotating ISS, point cloud registration, and classification tasks on $\mathfrak{sl}(3)$. These experiments clearly show the advantages of an adjoint-equivariant Lie algebraic network. We believe Lie Neurons could open new possibilities in both equivariant
modeling and more general deep learning on Lie algebras.

\section*{Impact Statement}
This paper presents work whose goal is to advance the field of Machine Learning. There are many potential societal consequences of our work, none which we feel must be specifically highlighted here.

\section*{Acknowledgments}
This work was supported by AFOSR MURI FA9550-23-1-0400.


{\small
\balance 
\bibliography{strings-full, ieee-full, main_icml}
\bibliographystyle{icml2024}
}
\newpage
\appendix
\onecolumn

\section{Additional Preliminaries}
\subsection{The Hat and Vee Operator}
\label{sec:hat_vee_example}
As introduced in Section~\ref{sec:preliminaries}, the Hat and Vee operators can be defined as~\cite{chirikjian2011stochastic}:

\begin{align}
    \text{Vee}: & \mathfrak{g} \rightarrow \mathbb{R}^m, \quad x^{\wedge} \mapsto (x^{\wedge})^\vee = \sum^m_{i=1} x_i e_i\label{eq:veehat}, \quad \nonumber\\
    \text{Hat}: & \mathbb{R}^m \rightarrow \mathfrak{g}, \quad x \mapsto x^{\wedge} = \sum^m_{i=1} x_i E_i, \nonumber
\end{align}
where $E_i = \in \mathfrak{g}$ are linear independent basis in $\mathfrak{g}$, and $e_i$ are the canonical basis of $\mathbb{R}^m$.

For example, the $\mathfrak{so}(3)$ elements are skew-symmetric, $\begin{bmatrix} 0& -\omega_z & \omega_y \\ \omega_z & 0 & -\omega_x \\ -\omega_y & \omega_x & 0 \end{bmatrix} \in \mathfrak{so}(3)$. One can find the canonical basis as follows:
\begin{equation}
\small 
E_x = \begin{bmatrix} 0& 0 & 0\\ 0&0&-1\\ 0&1&0 \end{bmatrix}, E_y = \begin{bmatrix} 0& 0 & 1\\ 0&0&0\\ -1&0&0 \end{bmatrix}, E_z = \begin{bmatrix} 0& -1 & 0 \\1&0&0\\ 0&0&0 \end{bmatrix}.
\label{eq:so3_basis}
\end{equation}

With $\begin{bmatrix} 0& -\omega_z & \omega_y \\ \omega_z & 0 & -\omega_x \\ -\omega_y & \omega_x & 0 \end{bmatrix} \in \mathfrak{so}(3)$, we have 
\begin{align}
\small 
    W &= \begin{bmatrix}0& -\omega_z & \omega_y \\ \omega_z & 0 & -\omega_x \\ -\omega_y & \omega_x & 0 \end{bmatrix}\\
     &= \omega_x E_x + \omega_y E_y + \omega_z E_z  \\
     &= v^\wedge,\\
    W^\vee &= v = \begin{bmatrix} \omega_x, \omega_y, \omega_z \end{bmatrix}^\transpose.
\end{align}
Using the Hat and Vee maps, we can represent an element of the Lie algebra in a neural network using $\mathbb{R}^m$, while performing structure-preserving operations on $\mathfrak{g}$. In this work, we use the basis of $\mathfrak{so}(3)$ in \eqref{eq:so3_basis}, and $\mathfrak{sl}(3)$ from~\cite{winternitz2004subalgebras} to construct the Hat and Vee maps.

\section{An Illustration of the Proposed Work}
\begin{figure*}
    \centering
    \includegraphics[width=0.75\textwidth]{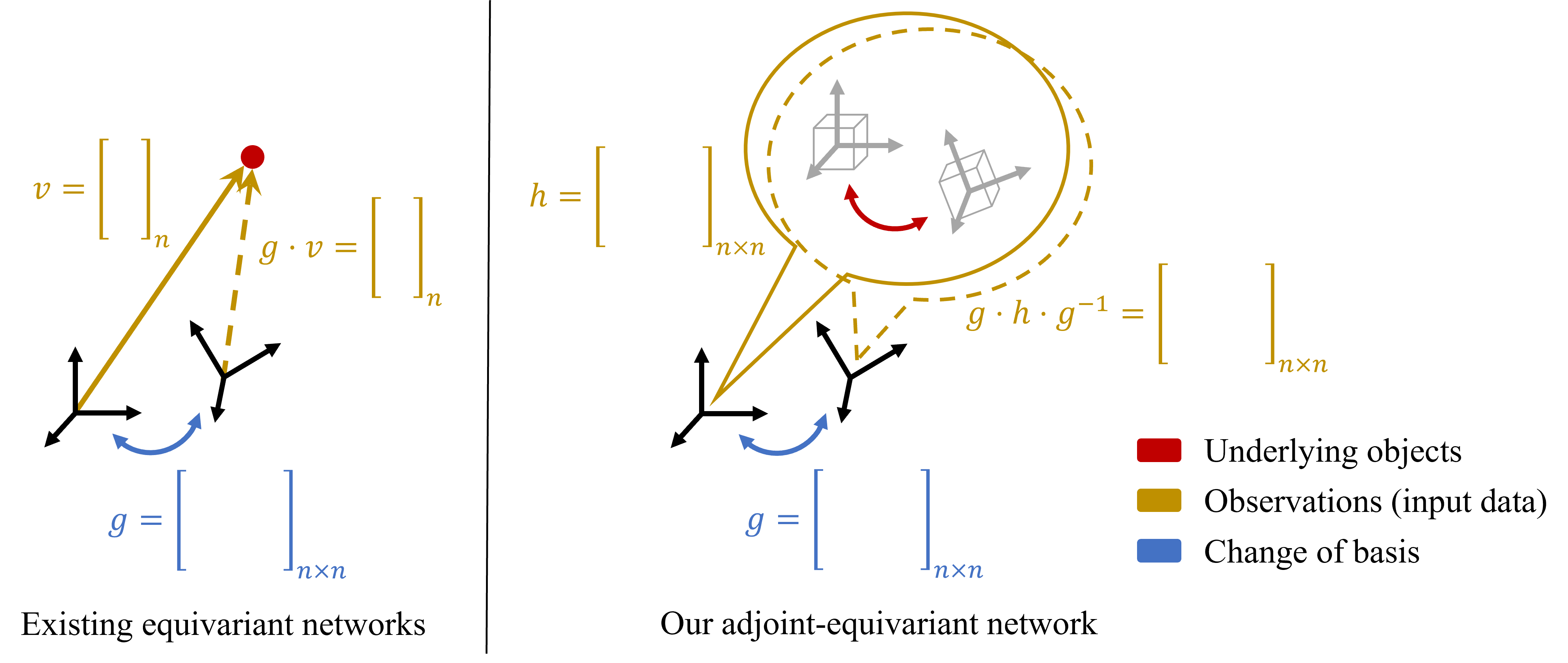}
    \caption{Comparison between existing equivariant networks and our work. The existing equivariant networks take in vectors in $\mathbb{R}^n$ and are equivariant to the left action of a group. Our adjoint equivariant network takes elements in the Lie algebra as inputs and is equivariant to the adjoint action, which corresponds to a change of basis operation.}
    \label{fig:title}
\end{figure*}

Figure \ref{fig:title} shows the comparison between existing equivariant networks and our work. The existing equivariant networks take in vectors in $\mathbb{R}^n$ and are equivariant to the left action of a group. Our adjoint equivariant network takes elements in the Lie algebra as inputs and is equivariant to the adjoint action, which corresponds to a change of basis operation.

\section{Implementation Details}
Figure \ref{fig:ln_layers} shows the different architecture used in each experiment. For the hyperparameter used in each experiment, we kindly refer the readers to the open-sourced GitHub repository: \hyperlink{https://github.com/UMich-CURLY/LieNeurons}{https://github.com/UMich-CURLY/LieNeurons}.

\subsection{Baker–Campbell–Hausdorff Formula}
The Lie Neurons structure used in the BCH experiments is shown in Figre~\ref{fig:ln_layers}. It consists of two \texttt{LN-LB+LN-LR} layers. The feature dimension of each layer is set to $1024$, while the last linear layer projects the features back to dimension $3$.

For Vector Neurons, we maintain the same architecture as Lie Neurons while changing all the \texttt{LN-LB} to the ReLU layer from VN. The feature dimension is also set to $1024$.

For EMLP, we construct 3 EMLP blocks, each consisting of a linear layer, a bilinear layer, and a gated nonlinearity. The channel size is set to 128 for each block. (We were unable to increase the feature dimension due to memory complexity.) The input representation is set to two $3 \times 1$ vectors with respect to $SO(3)$, and the output representation is set to one $3 \times 1$ vector with respect to $SO(3)$.

To construct an e3nn network for comparison, we use equivariant linear layers, batch normalization layers, and activation layers from the library. We experiment with both norm-based activation layers and spherical point-wise activation layers. We use type-0 to type-4 features in the hidden layers. We tune the hyperparameters with different choices of depth and feature dimensions. The reported result is using 6 linear-batchnorm-norm-based-activation layers. The network size is comparable to other networks in the comparison. In our experiment, the performance of e3nn is similar to Vector Neurons, both of which cannot successfully regress the BCH formula. It is possible that e3nn might be able to achieve better performance by designing more complicated architectures, nonlinearities, and further tuning the hyperparameters, but the experiment still confirms the superiority of our proposed LN model in this equivariant regression task on so(3) algebra.

\subsection{Dynamic Modeling}
When a change of reference frame is performed to a dynamical system, $\omega$ undergoes a left rotation action, while the inertia tensor undergoes a change of basis action. That is $R I R^{-1}$. The equivariance of the system can be shown using the following derivation:
\begin{align} f(\omega,I) &= - I^{-1} (\omega \times I \omega) = - I^{-1} [\omega]_\times I \omega, \ \end{align} where $[\cdot]_\times$ means skew symmetric form. Here we are using the fact that the cross product can be replaced with a skew symmetric matrix multiplication. By using $[R\omega]_\times = R[\omega]_\times R^{-1}$, we can have \begin{align} f(R\omega; RIR^{-1}) &= - RI^{-1}R^{-1} (R \omega) \times R I R^{-1} R \omega \end{align} \begin{align} &= - RI^{-1} R^{-1} R [\omega]_\times R^{-1} R I R^{-1} R \omega \end{align} \begin{align} &= - RI^{-1}[\omega]_\times I \omega = Rf(\omega,I) \end{align} This means that the symmetry of the system itself is with respect to both $\omega$ and $I$. If we force the network to learn the dynamics using only $\omega$ as the input, we cannot correctly capture the symmetry of the system. As a result, we introduce $m$ as steerable and learnable weights as additional input elements. During test time, the weights $m$ are fixed. When we rotate the system, we apply the same rotation on $m$. Since $m$ is introduced to address equivariance with respect to the inertia tensor, it can be viewed as an implicit representation of the inertia tensor. As shown in Figure~\ref{fig:ln_layers}, the learnable weights $m$ pass through two branches of \texttt{LN-LR} layers. After such, they are mixed with the input data $\textbf{x}$ in a bracket layer. For each layer, we set the feature dimension as $20$.

For the baseline EMLP~\cite{finzi2021practical}, we introduce the learnable weights $m$ as three $3 \times 1$ vectors to the system input to serve as the implicit representation of the inertia tensor. We construct 3 EMLP blocks, each consisting of a linear layer, a bilinear layer, and a gated nonlinearity. The channel size is set to 128 for each block. The input representation is set to four $3 \times 1$ vectors with respect to $SO(3)$ (a measurement $\mathbf{x}$ and three $3 \times 1$ vectors for $m$.), and the output representation is set to one $3 \times 1$ vector with respect to $SO(3)$.

\subsection{Point Cloud Registration}
We modified the open-sourced implementation provided in~\citet{zhu2022correspondence} for the point cloud registration task.~\footnote{We modified the open-sourced code from \hyperlink{https://github.com/minghanz/EquivReg}{https://github.com/minghanz/EquivReg}.} The inputs are two noisy point clouds with different orientations. The goal of the network is to regress an $SO(3)$ rotation that best aligns the two point clouds in a correspondence-free manner. We corrupt both the training and testing data using Gaussian noise with a standard deviation of $0.01$ after normalizing the point cloud to a unit cube. The initial rotation is sampled from (0,180) degrees randomly. Similar to~\citet{zhu2022correspondence}, we use PointNet as the encoder backbone for the network, and we replace the Vector Neurons modules with the Lie Neurons modules. In addition to Lie Neurons, we test the effect of the mixing layers with Vector Neurons. For that, we add the mixing layers before each ReLU in the original Vector Neurons while keeping the rest of the structure untouched. 

\subsection{Platonic Classification}
\begin{figure}[t]
    \centering
    \includegraphics[width=0.95\textwidth]{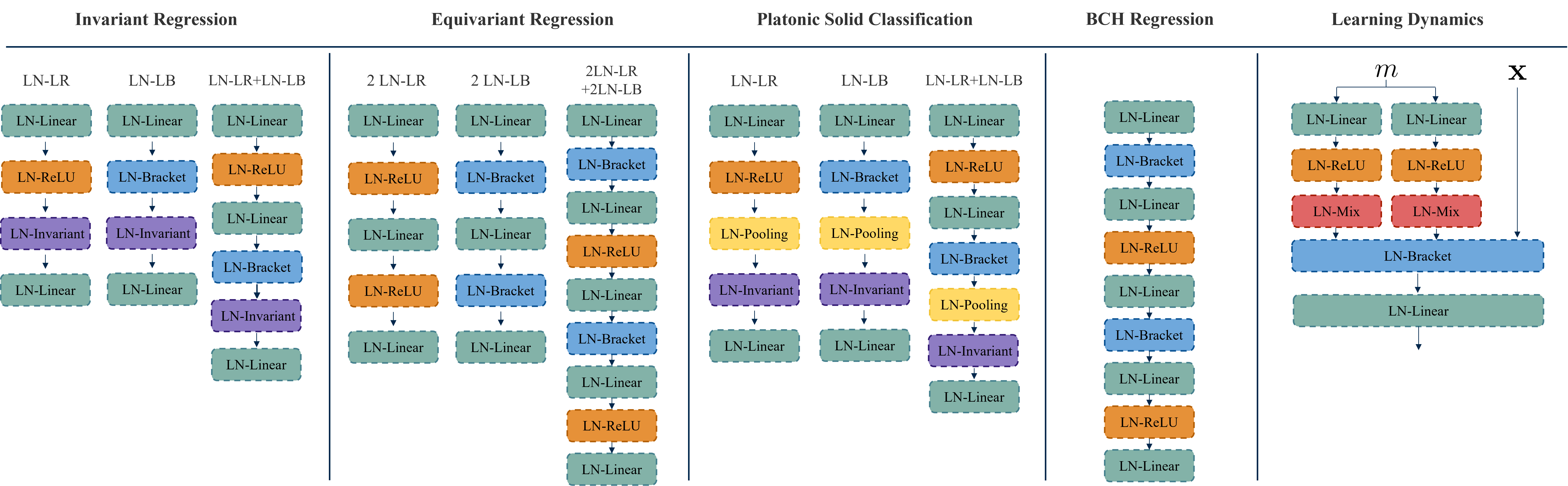}
    \caption{The network architecture used in each experiment.}
    \label{fig:ln_layers}
\end{figure}

The architectures used in this experiment are shown in Figure~\ref{fig:ln_layers}. Since classification is an invariant task, an invariant layer is attached to the end of each model. The feature dimension is set to 256 for each layer of the models. We compare our method with a standard 3-layer MLP, in which we also set the feature dimension to 256.

\section{Additional Experiments on $\mathfrak{sl}(3)$}
\label{sec:additional_sl3_exp}
We present two additional experiments on $\mathfrak{sl}(3)$, in which we ask the network to regress an invariant and an equivariant function. Across both experiments, we compare our method with a standard 3-layer MLP by flattening the input to $\mathbb{R}^{K\times C\times N}$. In addition, we set the feature dimension to 256 for all models.

\subsection{Invariant Function Regression}
\label{sec:sl3_inv}
We begin our evaluation with an invariant function fitting experiment. Given $X, Y \in \mathfrak{sl}(3)$, we ask the network to regress the following function:
\begin{equation}
\begin{split}
    g(X,Y) = & \sin(\tr(XY))+\cos(\tr(YY))-\frac{\tr(YY)^3}{2} \\
            & +\det(XY)+\exp(\tr(XX)).
\end{split}
\end{equation}
We randomly generate $10,000$ training samples and $10,000$ testing samples. In addition, in order to evaluate the invariance of the learned network, we randomly apply $500$ group adjoint actions to each test sample to generate augmented testing data.

In this task, we experiment with three different modules, \texttt{LN-LR}, \texttt{LN-LB}, and \texttt{LN-LR + LN-LB}, each followed by an \texttt{LN-Inv} and a final linear mapping from the feature dimension to a scalar. For each input, we concatenate $X$ and $Y$ in the feature dimension and have $\mathcal{X} \in \mathbb{R}^{K \times C \times N} = \mathbb{R}^{8 \times 2 \times 1}$. 
We additionally train the MLP with augmented data to serve as a stronger baseline.

To show the performance consistency, we train each model 5 separate times and calculate the mean and standard deviation of the performance. We report the Mean Squared Error (MSE) and the invariance error in Table~\ref{tab:inv_results}. The invariance error $E_{\text{inv}}$ is defined as:
\begin{equation}
    E_{\text{inv}} := \frac{\sum_{i=1}^{N_x}\sum_{j=1}^{N_a} f(\mathcal{X}_i) - f(a_j\mathcal{X}_ia_j^{-1})}{N_x N_a},
\end{equation}
where $a \in \mathrm{SL}(3)$ are the randomly generated adjoint actions, $N_x$ is the number of testing points, and $N_a$ is the number of conjugations. The invariance error measures the extent to which the model is invariant to the adjoint action.

From the table, we see that the LN outperforms MLP except for \texttt{LN-LB}. When tested on the $\mathrm{SL}(3)$ augmented test set, the performance of the LN remains consistent, while the error from the MLP increases significantly. The results of the invariance error demonstrate that the proposed method is invariant to the adjoint action while the MLP is not. Data augmentation helps MLP to perform better in the augmented test set, but at the cost of worse $Id$ test set performance, and the overall performance still lags behind our equivariant models. In this experiment, we observe that \texttt{LN-LR} performs well on the invariant task, but the \texttt{LN-LB} alone does not. Nevertheless, if we combine both nonlinearities, the performance remains competitive. 

We additionally provide the training curves in Figure~\ref{fig:training_curves} to analyze the convergence property of the proposed network. We can see that the proposed method converges faster than the MLP, which indicates it is more data efficient. In addition, the MLP overfits to the training set and underperforms on the test set, while our method remains consistent.
\begin{table*}[t]
    \centering
    \caption{The mean squared errors and the invariant errors on the $\mathfrak{sl}(3)$ invariant function regression task. $\downarrow$ means the lower the better.}
    \footnotesize
    \resizebox{\textwidth}{!}{
    \begin{tabular}{c|c|c|cc|cc|cc}
    \toprule
\multirow{3}{*}{Model} & \multirow{3}{*}{\begin{tabular}[c]{@{}c@{}}Training \\ Augmentation\end{tabular}} & \multirow{3}{*}{Num Params} & \multicolumn{4}{c}{Testing Augmentation}                       & \multicolumn{2}{|c}{\multirow{2}{*}{Equivariance Error}} \\ \cmidrule{4-7} 
                       &                                                                                   &                             & \multicolumn{2}{c}{$Id$}       & \multicolumn{2}{c|}{$\mathrm{SL}(3)$}    & \multicolumn{2}{c}{}                                  \\ \cmidrule{4-9}
                       &                                                                                   &                             & AVG  $\downarrow$             & STD      & AVG    $\downarrow$           & STD      & AVG  $\downarrow$                          & STD                  \\ \midrule
MLP                    & $Id$                                                                                & 136,193                     & 0.148             & 0.005    & 6.493             & 1.282    & 1.415                          & 0.113                \\
MLP                    & $\mathrm{SL}(3)$                                                                             & 136,193                     & 0.201             & 0.01     & 1.119             & 0.018    & 0.683                          & 0.006                \\
LN-LR                  & $Id$                                                                                & 66,562                      & 1.30 $\times$ 10$^{-3}$          & 3.24 $\times$ 10$^{-5}$ & 1.30 $\times$ 10$^{-3}$          & 3.25 $\times$10$^{-5}$ & 3.60 $\times$ 10$^{-4}$              & 5.48 $\times$ 10$^{-5}$             \\
LN-LB                  & $Id$                                                                                & 132,098                     & 0.557             & 1.87 $\times$ 10$^{-4}$ & 0.557             & 1.87 $\times$ 10$^{-4}$ & $\mathbf{1.43\times10^{-5}}$                              & 1.42$\times$10$^{-6}$             \\
LN-LR + LN-LB          & $Id$                                                                                & 263,170                     & $\mathbf{8.84\times 10^{-4}}$ & 2.52$\times$10$^{-5}$ & $\mathbf{8.84\times10^{-4}}$ & 2.49$\times$10$^{-5}$ & 4.00$\times$10$^{-4}$                       & 0                   \\
\bottomrule
\end{tabular}
}
    \label{tab:inv_results}
    \vspace{-2mm}
\end{table*}

\subsection{Equivariant Function Regression}
In the second experiment, we ask the network to fit an equivariant function that takes two elements on $\mathfrak{sl}(3)$ back to itself:
\begin{align}
    h(X,Y) &= \left[\left[X,Y\right],Y\right] + \left[Y,X\right].
\end{align}
Similar to the first experiment, we generate $10,000$ training and test samples, as well as the additional $500$ adjoint actions on the test set. For this task, we also train each model separately 5 times to analyze the consistency of the proposed method. We again report the MSE on the regular test set. For the adjoint-augmented test set, we map the output back with the inverse adjoint action and compute the MSE with the ground truth value. To evaluate the equivariance of the network, we compute the equivariance error $E_{\text{equiv}}$ as:
\begin{equation}
    E_{\text{equiv}} := \frac{\sum_{i=1}^{N_x}\sum_{j=1}^{N_a} a_jf(\mathcal{X}_i)a_j^{-1} - f(a_j\mathcal{X}_ia_j^{-1})}{N_x N_a}.
\end{equation}
In this experiment, we evaluate LN using 3 different architectures. They are 2 \texttt{LN-LR}, 2 \texttt{LN-LB}, and 2 \texttt{LN-LR} + 2 \texttt{LN-LB}, respectively. Each of them is followed by a regular linear layer to map the feature dimension back to $1$. 

Table~\ref{tab:equiv_results} lists the results of the equivariant experiment. We see that the MLP performs well on the regular test set but fails to generalize to the augmented data. Moreover, it has a high equivariance error. Similar to the invariant task, data augmentation improves the MLP's performance on the augmented test set, but at the cost of worse $Id$ test set performance, and the overall performance still lags behind our equivariant models. Our methods, on the other hand, generalize well on the adjoint-augmented data and achieve the lowest errors. The 2 \texttt{LN-LB} model performs the best. 

The training curves of this experiment is shown in Figure~\ref{fig:training_curves}. The proposed network converges much faster than the MLP, which again demonstrates the data efficiency of the equivariant method.

From both the invariant and equivariant experiments, we observe that the \texttt{LN-LR} module works better on invariant tasks, while the \texttt{LN-LB} module performs better on the equivariant ones. We speculate this is because the \texttt{LN-LR} relies on the Killing form, which is an adjoint-invariant function, while the \texttt{LN-LB} leverages the Lie bracket, which is adjoint-equivariant. Nevertheless, if we combine both modules, the network performs favorably on both invariant and equivariant tasks. 

\begin{table*}[t]
    \centering
    \caption{The mean squared errors and the equivariant errors in $\mathfrak{sl}(3)$ equivariant function regression.}
    \footnotesize
    \resizebox{\textwidth}{!}{
    \begin{tabular}{c|c|c|cc|cc|cc}
    \toprule
\multirow{3}{*}{Model} & \multirow{3}{*}{\begin{tabular}[c]{@{}c@{}}Training \\ Augmentation\end{tabular}} & \multirow{3}{*}{Num Params} & \multicolumn{4}{c}{Testing Augmentation}                       & \multicolumn{2}{|c}{\multirow{2}{*}{Invariance Error}} \\ \cmidrule{4-7} 
                       &                                                                                   &                             & \multicolumn{2}{c}{$Id$}       & \multicolumn{2}{c|}{$\mathrm{SL}(3)$}    & \multicolumn{2}{c}{}                                  \\ \cmidrule{4-9}
                       &                                                                                   &                             & AVG     $\downarrow$          & STD      & AVG    $\downarrow$           & STD      & AVG    $\downarrow$                        & STD                  \\ \midrule
MLP                    & $Id$                                                         & 538,120                     & 0.011      & 3.53$\times$10$^{-4}$   & 1.318       & 7.08$\times$10$^{-2}$    & 0.424                      & 0.003                      \\
MLP                    & $\mathrm{SL}(3)$                                                      & 538,120                     & 0.033      & 2.86$\times$10$^{-4}$   & 0.452       & 1.01$\times$10$^{-2}$    & 0.389                      & 0.001                      \\
2 LN-LR                & $Id$                                                         & 197,376                     & 0.213    & 4.07$\times$10$^{-5}$   & 0.213       & 4.08$\times$10$^{-5}$    & 9.32$\times$10$^{-5}$                   & 6.65$\times$10$^{-6}$                   \\
2 LN-LB                & $Id$                                                         & 328,448                     & $\mathbf{9.83\times10^{-10}}$   & 1.78$\times$10$^{-11}$   & $\mathbf{4.55\times10^{-8}}$    & 8.65$\times$10$^{-11}$    & $\mathbf{6.56\times10^{-5}}$                   & 4.22$\times$10$^{-7}$                   \\
2 LN-LR + 2 LN-LB      & $Id$                                                         & 590,592                     & 7.65$\times$10$^{-9}$   & 3.54$\times$10$^{-10}$   & 5.41$\times$10$^{-8}$    & 4.08$\times$10$^{-10}$   & 7.67$\times$10$^{-5}$                   & 1.56$\times$10$^{-6}$                  \\ \bottomrule
\end{tabular}}
    \label{tab:equiv_results}
    \vspace{-2mm}
\end{table*}

\section{Ablation Study on Skipping Connection in the Bracket Layer}
We introduce the \texttt{LN-Bracket} layer in \cref{sec:ln-bracket} and discuss how the residual connection improves the performance. In this subsection, we perform ablation studies on an alternative Lie bracket nonlinear layer design without the residual connection. That is,
$
    f_{\text{LN-Bracket-N}}(\mathbf{x})= [(\mathbf{x}U)^\wedge, (\mathbf{x}V)^\wedge]^\vee
$. 
We denote this nonlinear layer combined with an \texttt{LN-Linear} as \texttt{LN-LBN} and show the results of this method in Table~\ref{tab:ablation_results}. From the table, we can clearly see the benefits of having the residual connection in the Lie bracket layer.
 
\begin{table*}[t]
    \vspace{-5pt}
    \centering
    \footnotesize
    \caption{The ablation study of the Lie bracket layer in all three tasks. \texttt{LN-LBN} denotes the Lie bracket layer without the residual connection.}
    \small
    \resizebox{\textwidth}{!}{
    \begin{tabular}{l|c c c|c c c|c c}
    \toprule
        & \multicolumn{3}{c|}{$\mathfrak{sl}(3)$ Invariant Regression} & \multicolumn{3}{c|}{$\mathfrak{sl}(3)$ Equivariant Regression} & \multicolumn{2}{c}{Platonic Solid -Classification}\\
         \midrule
         & MSE $\downarrow$ & MSE $\mathrm{SL}(3) \downarrow$ & $E_{\text{inv}} \downarrow$ & MSE $\downarrow$ & MSE $\mathrm{SL}(3) \downarrow$ & $E_{\text{equiv}} \downarrow$ & Acc $\uparrow$ & Acc (Rotated) $\uparrow$ \\
         \hline
         \texttt{LN-LB} & \textbf{0.558} & \textbf{0.558} & $4.9 \times 10^{-5}$ & $\mathbf{9.6 \times 10^{-10}}$ & $\mathbf{4.5 \times 10^{-8}}$ & $\mathbf{6.5 \times 10^{-5}}$ & \textbf{0.986} & \textbf{0.979}\\
        \texttt{LN-LBN}  & 4.838 & 4.838 & $\mathbf{2.4 \times 10^{-5}}$ & 0.276 & 0.276 & $2.7 \times 10^{-3}$ & 0.967 & 0.959\\
        \bottomrule
    \end{tabular}
    }
    \label{tab:ablation_results}
    \vspace{-2mm}
\end{table*}

\section{Training Curves}
\begin{figure*}[h]
    \centering
    \includegraphics[width=0.45\textwidth]{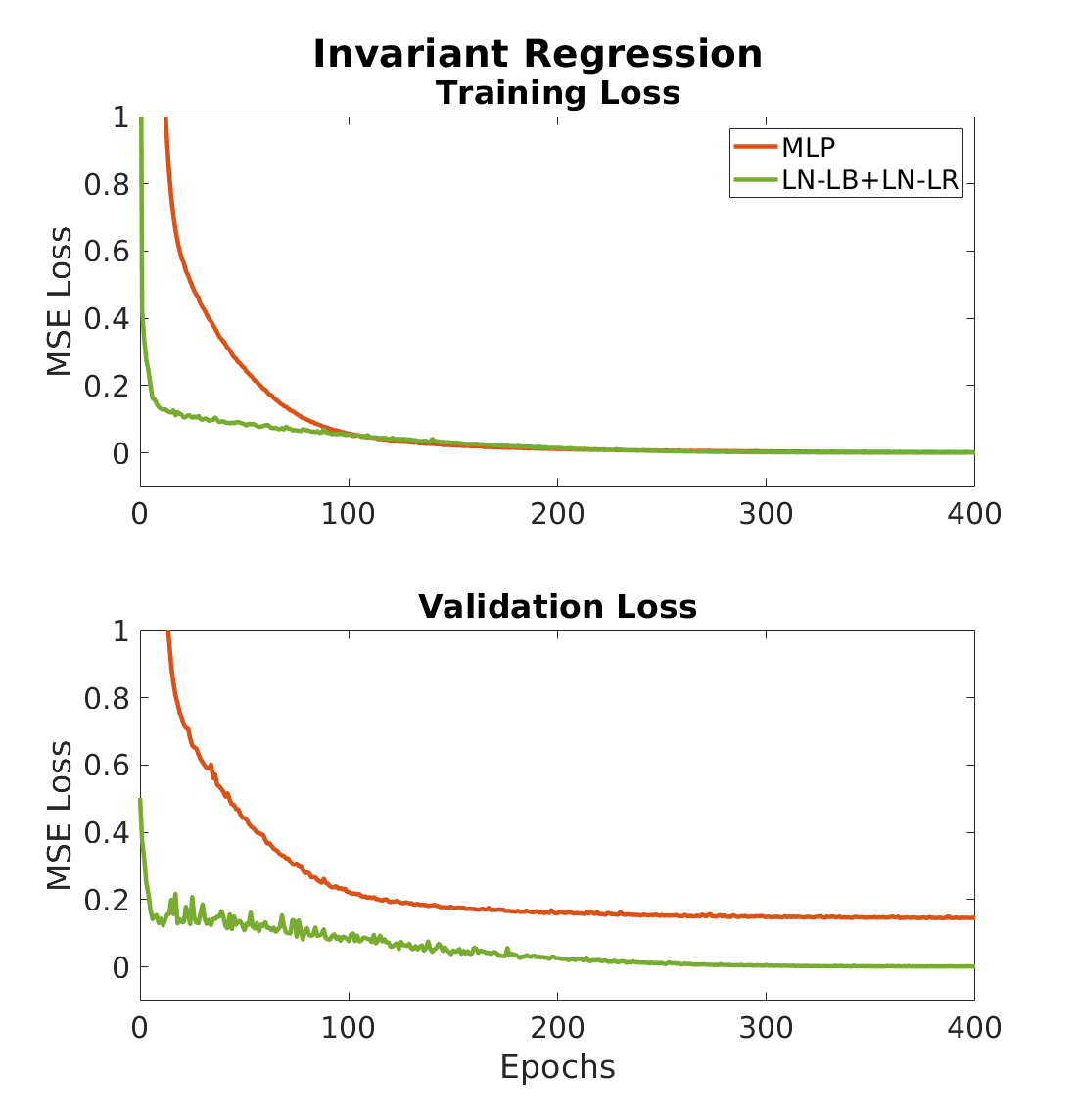}
    \includegraphics[width=0.45\textwidth]{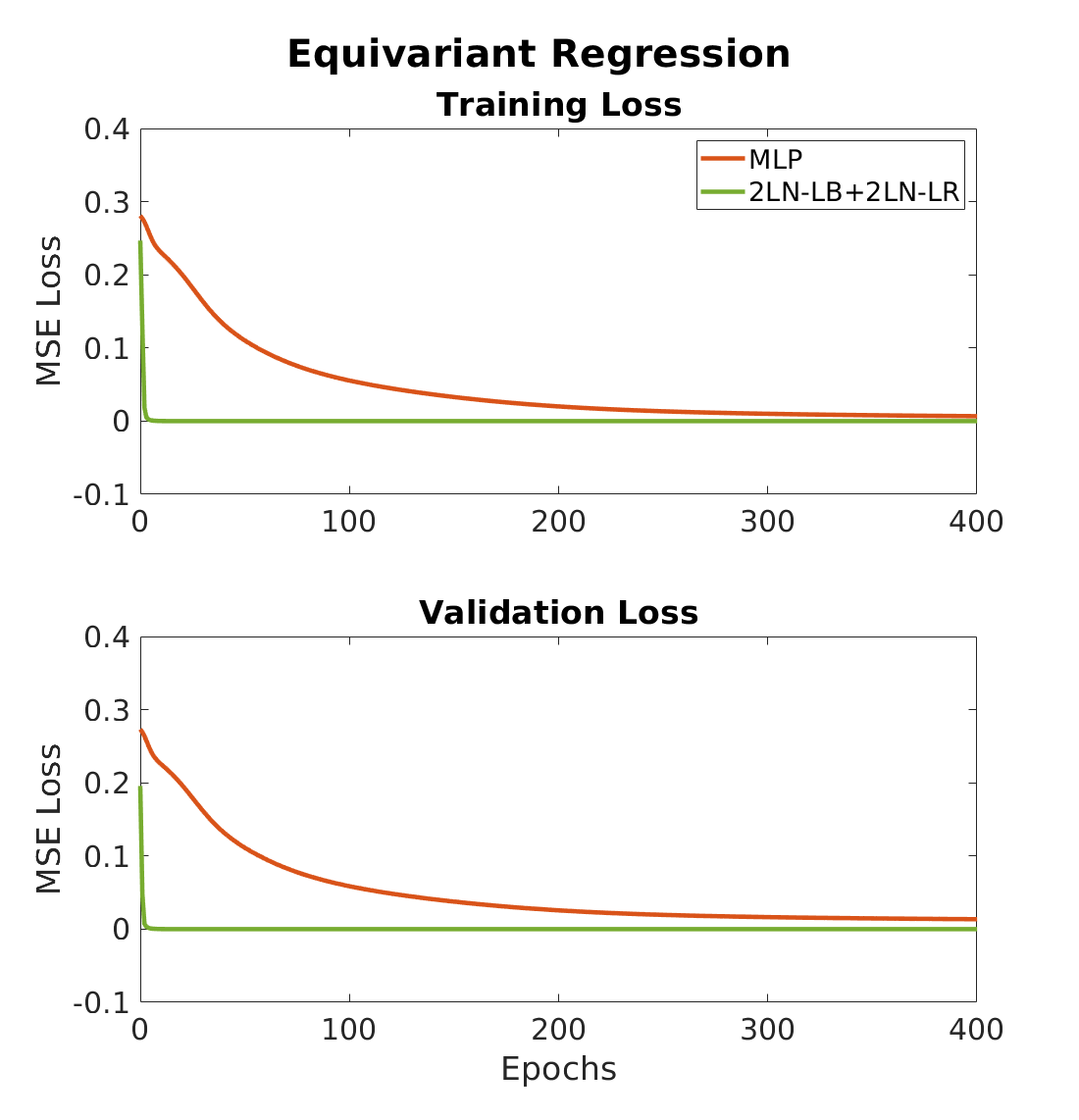}
     \caption{The training curves of the MLP and the proposed method. We can see that in both invariant and equivariant regression tasks, our equivariant model converges much faster than MLPs, showing the data efficiency of our method.}
     \label{fig:training_curves}
\end{figure*}

We present the training curves for the $\mathfrak{sl}(3)$ invariance and equivariance regression tasks in Figure~\ref{fig:training_curves}. In both tasks, LN converges faster than MLPs, showing the data efficiency of the proposed method.


\end{document}